\documentclass[sigconf]{acmart}

\usepackage{makecell}
\usepackage{cleveref}
\usepackage{amsmath}

\AtBeginDocument{%
  }

\setcopyright{acmlicensed}
\copyrightyear{2025} 
\acmYear{2025} 
\setcopyright{acmlicensed}\acmConference[MM '25]{Proceedings of the 33rd
ACM International Conference on Multimedia}{October 27--31, 2025}{Dublin,
Ireland}
\acmBooktitle{Proceedings of the 33rd ACM International Conference on
Multimedia (MM '25), October 27--31, 2025, Dublin, Ireland}
\acmDOI{10.1145/3746027.3755220}
\acmISBN{979-8-4007-2035-2/2025/10}

\begin{document}

\title{Inversion-DPO: Precise and Efficient Post-Training for Diffusion Models}

\author{Zejian Li}
\authornote{Corresponding author}
\orcid{0000-0001-5313-2742}
\affiliation{%
  \institution{Zhejiang University}
  %\department{School of Software Technology}
  \city{Ningbo}
  %\state{Zhengjiang}
  \postcode{315048}
  \country{China}
}
\email{zejianlee@zju.edu.cn}

\author{Yize Li}
\orcid{0009-0006-6774-6319}
\affiliation{%
  \institution{Zhejiang University}
  %\department{School of Software Technology}
  \city{Ningbo}
  %\state{Zhengjiang}
  \postcode{315048}
  \country{China}
}
\email{1101200212@stu.jiangnan.edu.cn}

\author{Chenye Meng}
\orcid{0000-0002-4787-6232}
\affiliation{%
  \institution{Zhejiang University}
  %\department{College of Computer Science and Technology}
  \city{Hangzhou}
  %\state{Zhengjiang}
  \postcode{310027}
  \country{China}
}
\email{mengcy@zju.edu.cn}

\author{Zhongni Liu}
\orcid{0009-0007-8449-1802}
\affiliation{%
    \institution{University of Electronic Science and Technology of China}
   % \department{School of Information and Software Engineering}
    \city{Chengdu}
    %\state{Sichuan}
    \postcode{610054}
    \country{China}
}
\email{2022091203026@std.uestc.edu.cn}

\author{Ling Yang}
\orcid{0000-0003-1905-8053}
\affiliation{%
  \institution{Peking University}
  \city{Beijing}
  \country{China}}
\email{yangling0818@163.com}

\author{Shengyuan Zhang}
\orcid{0000-0003-3762-1612}
\affiliation{%
  \institution{Zhejiang University}
  %\department{College of Computer Science and Technology}
  \city{Hangzhou}
 %\state{Zhengjiang}
  \postcode{310027}
  \country{China}
}
\email{zhangshengyuan@zju.edu.cn}

\author{Guang Yang}
\orcid{0000-0001-8061-742X}
\affiliation{%
  \institution{Alibaba Group}
 % \department{Design}
  \city{Hangzhou}
  %\state{Zhengjiang}
  \postcode{310000}
  \country{China}
}
\email{qingyun@alibaba-inc.com}

\author{Changyuan Yang}
\orcid{0000-0003-0065-6272}
\affiliation{%
  \institution{Alibaba Group}
 % \department{Design}
  \city{Hangzhou}
  %\state{Zhengjiang}
  \postcode{310000}
  \country{China}
}
\email{changyuan.yangcy@alibaba-inc.com}

\author{Zhiyuan Yang}
\orcid{0009-0005-4950-671X}
\affiliation{%
  \institution{Alibaba Group}
 % \department{Design}
  \city{Hangzhou}
  %\state{Zhengjiang}
  \postcode{310000}
  \country{China}
}
\email{adam.yzy@alibaba-inc.com}

\author{Lingyun Sun}
\orcid{0000-0002-5561-0493}
\affiliation{%
  \institution{Zhejiang University}
 % \department{College of Computer Science and Technology}
  \city{Hangzhou}
  %\state{Zhengjiang}
  \postcode{310027}
  \country{China}
}
\email{sunly@zju.edu.cn}

\renewcommand{\shortauthors}{Zejian Li et al.}

\begin{abstract}
Recent advancements in diffusion models (DMs) have been propelled by alignment methods that post-train models to better conform to human preferences. 
However, these approaches typically require computation-intensive training of a base model and a reward model, which not only incurs substantial computational overhead but may also compromise model accuracy and training efficiency. 
To address these limitations, we propose Inversion-DPO, a novel alignment framework that circumvents reward modeling by reformulating Direct Preference Optimization (DPO) with DDIM inversion for DMs. 
Our method conducts intractable posterior sampling in Diffusion-DPO with the deterministic inversion from winning and losing samples to noise and thus derive a new post-training paradigm. 
This paradigm eliminates the need for auxiliary reward models or inaccurate appromixation, significantly enhancing both precision and efficiency of training.
We apply Inversion-DPO to a basic task of text-to-image generation and a challenging task of compositional image generation.
Extensive experiments show substantial performance improvements achieved by Inversion-DPO compared to existing post-training methods and highlight the ability of the trained generative models to generate high-fidelity compositionally coherent images. 
For the post-training of compostitional image geneation, we curate a paired dataset consisting of 11,140 images with complex structural annotations and comprehensive scores, designed to enhance the compositional capabilities of generative models.
Inversion-DPO explores a new avenue for efficient, high-precision alignment in diffusion models, advancing their applicability to complex realistic generation tasks.
Our code is available at \url{https://github.com/MIGHTYEZ/Inversion-DPO}
\end{abstract}

\begin{CCSXML}
<ccs2012>
<concept>
<concept_id>10010147.10010178.10010224</concept_id>
<concept_desc>Computing methodologies~Computer vision</concept_desc>
<concept_significance>500</concept_significance>
</concept>
</ccs2012>
\end{CCSXML}

\ccsdesc[500]{Computing methodologies~Computer vision}

\keywords{Diffusion models, Direct Preference Optimization, Post-training}


\maketitle

%前面主要讲fundamental，一系列task放到一两句话。在一系列任务验证，t2i和更复杂的任务。
%弱化LAION-SG，说造了一个pair的数据，不要提具体数据集，可以在实现细节讲一下用了LAION-SG
%SG不要讲，只在数据集实现细节的时候说一两句，其他都用structure
%intro（分两段）和method把题目两点强调说，实验也是从这两个点。
%引用全面，其他文章有的指标都放上

\section{Introduction}

\begin{figure*}[t]
    \centering
    \includegraphics[width=1\textwidth]{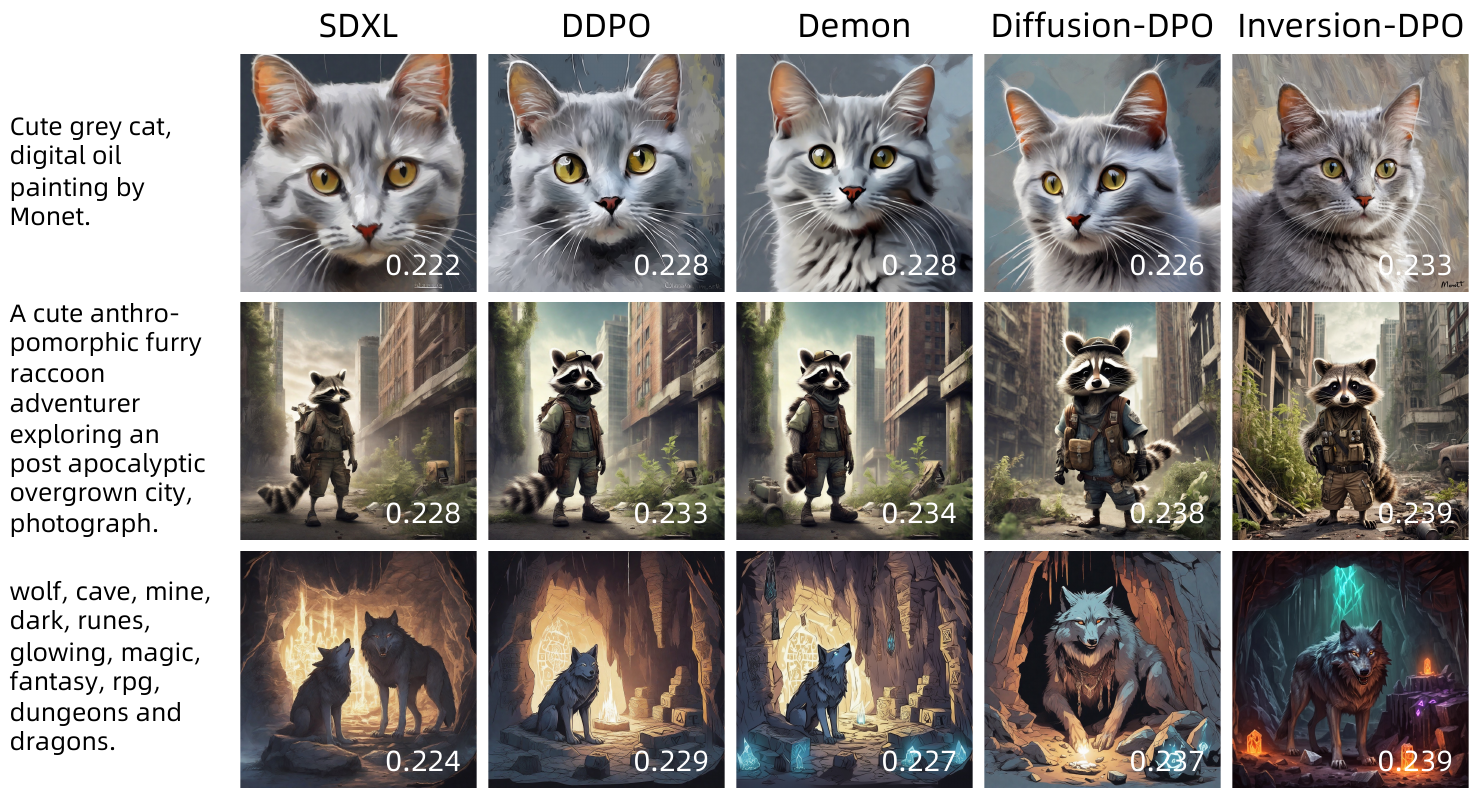} % Reduce the figure size so that it is slightly narrower than the column.
    % \vspace{-1.5em}
    \caption{Visual comparison of different baselines and Inversion-DPO. 
    % Each row presents an input prompt and the corresponding images generated by different models. 
    Each image is annotated with PickScore$^\uparrow$ at the bottom right. Inversion-DPO outperforms the baselines in both human visual perception and metric evaluation.
    }
    \label{fig: teaser}
    % \vspace{-1em}
\end{figure*}

Diffusion models (DMs)~\cite{saharia2024Photorealistic_imagen, dalle2, DM_beats_GAN, Chen2024PixArtWT, Tewel2024TrainingFreeCT, Shifted_Diffusion, Generative_Image_Dynamics, podell2023sdxlimprovinglatentdiffusion, Jiang2024Diffmm} have achieved state-of-the-art performance in image generation. This advancement is largely attributed to recent developments in alignment methods of DMs~\cite{jiang2024realigndiff,Uehara2024Feedback,li2024reward,Domingo2025adjoint,Zhang2024aligning,xu2024ImageReward, Li2024ClickDiff}, which enable models to be fine-tuned to better align with specific standards or human preferences. These methods typically require training two models simultaneously: a base model and a reward model. While this training paradigm incentivizes superior model performance, it also introduces significant computational overhead.

Diffusion-DPO~\cite{Wallace2024DiffusionDPO} reparameterizes the reward score into the loss function of the base diffusion model, eliminating the need for an auxiliary model and establishing a new paradigm for DMs to learn from human preferences. However, we identify a post-training issue in Diffusion-DPO, primarily due to its approximation of the reverse process using the forward distribution during derivation. This approximation leads to suboptimal computational accuracy and inefficient training.

% To overcome this limitation, we draw insights from DDIM inversion~\cite{song2020ddim} and propose Inversion-DPO, a general and efficient fine-tuning method for diffusion models. Inversion-DPO integrates the DDIM inversion strategy, transforming previously approximated terms--due to inaccessibility--into directly computable ones. This approach not only enhances computational accuracy but also simplifies redundant approximations, significantly improving both training efficiency and precision.

To overcome this limitation, we draw insights from DDIM inversion~\cite{song2020ddim} to establish a precise alignment framework for diffusion model fine-tuning. Our Inversion-DPO method fundamentally enhances approximation accuracy by adopting DDIM Inversion to approximate the reverse process \( p_\theta(x_{1:T} | x_0) \). This proposal eliminates the distributional mismatch inherent in previous approaches, improving approximation accuracy of sampling trajectories reversely from a sample $x_0$ to inversion samples $x_{1:T}$.

Such precision further translates to substantial efficiency gains through simplification of the final loss. By leveraging the trajectory of DDIM Inversion, our method reduces the original four KL divergence terms in Diffusion-DPO's derivation to two matching terms between aligned model predictions and reference model prediction. 
This enables more than 2$\times$ faster training convergence compared to baseline implementations. 
%上面这句话需要修改成具体数据，根据效率图
The efficiency improvements are particularly crucial for large-scale diffusion models like SDXL~\cite{podell2023sdxlimprovinglatentdiffusion}, where traditional approximation methods incur high computational costs.

To achieve high-quality and reliable image generation, we generalize Inversion-DPO to two specific tasks: basic text-to-image generation and compositional image generation. Extensive experiments demonstrate the effectiveness of our method in both diffusion model training and inference, as well as its superior performance compared with current SOTA methods, achieving significant improvements in key evaluation metrics (mention the specific improvements, e.g., percentage gains).
\Cref{fig: teaser} presents representative generated samples.
Furthermore, we introduce a novel dataset with structural annotations and complex scene preference scores, enabling structured evaluation and deeper insight into Inversion-DPO’s compositional preference learning.

In summary, by leveraging the principles of DDIM inversion, we address the computational accuracy and efficiency limitations of DPO-based methods. The optimized diffusion model achieves state-of-the-art performance in both basic text-to-image and compositional image generation tasks.

\section{Related Work}
\textbf{Alignment of Diffusion Models.} As generative models are widely used, aligning outputs with user preferences has become a key focus. Researchers are integrating Reinforcement Learning from Human Feedback (RLHF)~\cite{Ouyang2022Training,bai2022traininghelpfulharmlessassistant} into diffusion models to improve controllability and accuracy. 
A common approach involves using pre-trained or additional reward models~\cite{jiang2024realigndiff,Uehara2024Feedback,li2024reward,Domingo2025adjoint,Zhang2024aligning}, leveraging human feedback and external signals to fine-tune generation. ImageReward~\cite{xu2024ImageReward} trains a reward model based on human feedback to improve image quality and semantic alignment. DPOK~\cite{Fan2023DPOK} and DDPO~\cite{Black2024DDPO} combine reward signals with reinforcement learning for fine-tuning diffusion models, guiding generation through real-time feedback. ReNO~\cite{Eyring2024Reno} introduces a strategy that adjusts initial noise with reward signals, bypassing model training. IterComp~\cite{zhang2024itercomp} collects and combines model preferences, using iterative feedback to enhance generative capability.

The methods above typically require explicit training of reward models. Alternatively, some approaches bypass reward models and directly fine-tune generative models via reinforcement learning~\cite{Majumder2024Tango}. For example, Diffusion-DPO~\cite{Wallace2024DiffusionDPO} simplifies training by using differentiable reward signals for end-to-end fine-tuning. D3PO~\cite{Yang2024D3PO} eliminates explicit reward model training by using human feedback to guide the process. A Dense Reward View~\cite{yang2024adensereward} enhances reward fine-tuning with dense feedback at each denoising step, improving efficiency and stability. SPO~\cite{Liang2025SPO} adjusts denoising performance at each step with a preference model and resampler. While these DPO-based methods reduce reward model overhead, they still face post-training challenges. 
Recent works have explored using inversion techniques to improve DPO for diffusion models. DDIM-InPO~\cite{lu2025inpoinversionpreferenceoptimization} employs a reparameterization approach that treats diffusion models as single-step generators, selectively fine-tuning only the latent variables most correlated with preference data. SmPO-Diffusion~\cite{lu2025smoothedpreferenceoptimizationrenoise}introduces smoothed preference distributions combined with a Renoise Inversion technique, performing multiple DDIM inversion steps followed by an additional renoise step to handle preference variability. 
%In contrast, our Inversion-DPO uses DDIM inversion to reformulate DPO on the diffusion path, improving accuracy and efficiency.
%In contrast, our Inversion-DPO directly leverages complete DDIM inversion to precisely compute the intractable posterior $p_\theta(x_{1:T}|x_0)$, eliminating the approximation errors inherent in previous methods and achieving a more efficient loss formulation with only two matching terms instead of four KL divergences, improving accuracy and efficiency.
Our Inversion-DPO also leverages complete DDIM inversion. Out attempt is to precisely compute the intractable posterior $p_\theta(x_{1:T}|x_0)$, eliminating the approxima- tion errors inherent in previous methods and achieving a more efficient loss formulation. Our derivation results in only two matching terms instead of four KL divergences in Diffusion-DPO, improving accuracy and efficiency.

\textbf{Compositional Image Generation.} Compositional Text-to-Image Generation is a challenge, especially when involving multiple objects and complex relationships~\cite{yang2024mastering,Huang2024StableMoFusion}. While many studies have improved generative models, the sequential text format still limits results, particularly in compositional image generation. Researchers have proposed methods to enhance control and spatial awareness. For instance, Compositional Diffusion~\cite{Compositional_Diffusion} and Attend-and-Excite~\cite{Attend-and-Excite} improve generation efficiency but face limitations in complex scenes. Methods like GLIGEN~\cite{Li2023gligen} and Ranni~\cite{Feng2024Ranni} incorporate spatial conditioning and multimodal information to improve control, though they come with high training costs. BoxDiff~\cite{Xie2023BoxDiff} and RealCompo~\cite{zhang2024realcompo} optimize cross-attention and balance realism and complexity but remain dependent on bounding box accuracy. MIGC~\cite{zhou2024migc} and MIGC++~\cite{zhou2024migc++} address multi-instance composition using multimodal descriptions but focus on spatial control, failing to fully address abstract semantic relationships between objects.

To overcome the limitations of text formats, some studies have focused on using scene graphs for compositional generation~\cite{johnson2018image,ashual2019specifying,Compositional_Diffusion,yang2024mastering,Wang2023Symmetrical,Han2024Scenediffusion}. Scene graphs (SG) consist of nodes and edges representing objects and their relationships. Compositional SG2IM methods aim to generate high-quality images with multiple objects and complex relationships by better understanding and combining scene elements~\cite{feng2023trainingfree,Wang2024compositional,Wu2023Scene}.

SGDiff~\cite{Yang2022DiffusionBasedSG} enhances scene graph-based image generation by pretraining a scene graph encoder and integrating it with Stable Diffusion. SG-Adapter~\cite{Shen2024SGAdapterET} fine-tunes Stable Diffusion to incorporate scene graph information, improving image quality and semantic consistency. R3CD~\cite{Liu2024R3CD} introduces SG Transformers to expand diffusion models, learning abstract object interactions in larger datasets. DisCo~\cite{wang2024scene} combines scene graph decomposition with VAEs and diffusion models for more diverse outputs. The LAION-SG dataset~\cite{Li2024LAIONSG} further improves the model’s understanding of complex scenes.

Despite progress, compositional image generation still faces challenges in fidelity and efficiency. Our proposed method shows superior performance on difficult downstream tasks.

\section {Preliminary}

\subsection{Diffusion-DPO}
Reinforcement Learning from Human Feedback (RLHF) typically involves two stages: training a reward model from pairwise preferences and optimizing policies via reinforcement learning (RL). Direct Preference Optimization (DPO)~\cite{rafailov2023directDPO} bypasses this complexity by directly optimizing policies using preference data under a classification objective. 
Formally, given preference pairs \((x_0^w, x_0^l)\) conditioned on prompts \(c\), the Bradley-Terry model defines preference likelihood as:
\begin{equation}
    p_{\text{BT}}(x_0^w \succ x_0^l | c) = \sigma\left(r(c, x_0^w) - r(c, x_0^l)\right),
\end{equation}

Here \(r(\cdot, \cdot)\) is a reward function. Traditional RLHF maximizes the KL-regularized reward:
\begin{equation}
    \max_{p_\theta} \mathbb{E}_{x_0 \sim p_\theta(x_0|c)}[r(c,x_0)] - \beta \mathbb{D}_{\text{KL}}[p_\theta(x_0|c) \| p_{\theta_0}(x_0|c)],
\end{equation}
with \(p_{\theta}\) as the distribution to be optimized and \(p_{\theta_0}\) a reference distribution. DPO reparameterizes the optimal policy \(p_\theta^*\) as:
\begin{equation}
    p_\theta^*(x_0|c) \propto p_{\theta_0}(x_0|c) \exp\left(r(c,x_0)/\beta\right).
\end{equation}
Substituting this into the Bradley-Terry likelihood eliminates the reward function, yielding the DPO loss:
\begin{equation}
    \mathcal{L}_{\text{DPO}} = -\mathbb{E}_{(c,x_0^w,x_0^l)} \left[ \log \sigma \left( \beta \log \frac{p_\theta(x_0^w|c)}{p_{\theta_0}(x_0^w|c)} - \beta \log \frac{p_\theta(x_0^l|c)}{p_{\theta_0}(x_0^l|c)} \right) \right].
\end{equation}
This implicitly optimizes rewards while maintaining policy proximity to \(p_{\theta_0}\), avoiding explicit reward modeling and RL instability.

% \textbf{Diffusion-DPO.}
For diffusion models, the challenge lies in defining likelihoods over generated images \(x_0\). Diffusion-DPO~\cite{Wallace2024DiffusionDPO} extends DPO by leveraging the evidence lower bound (ELBO) of diffusion processes. 
The KL-regularized reward maximization objective becomes:
\begin{equation}
\label{eq:reward_maximization}
    \max_{p_\theta} \mathbb{E}_{p_\theta(x_{1:T}|c, x_0)} \left[ R(c, x_{0:T}) \right] - \beta \mathbb{D}_{\text{KL}}[p_\theta(x_{0:T}|c) \| p_{\theta_0}(x_{0:T}|c)],
\end{equation}
Here \(R(c, x_{0:T})\) denotes the reward over the full diffusion trajectory. 
Given a wining or losing sample $x_0$, a posterior sampling \(p_\theta(x_{1:T}|x_0, c)\) is required to gain the noisy sample sequence \(x_{1:T}\) of $p_\theta$, which is intractable.
% generates images through a Markov chain of latent variables \(x_{1:T}\). 
By approximating the reverse process with the forward process \(q(x_{1:T}|x_0)\), the loss simplifies to:
\begin{align}
\label{eq:diffusion-DPO-final}
    \mathcal{L}_{\text{Diffusion-DPO}} = -\mathbb{E}_{(x_0^w, x_0^l), t, q(x_t^w\mid x_0^w), q(x_t^l\mid x_0^l)} \log \sigma ( -\beta T \omega(\lambda_t) ( \notag \\
    \| \epsilon^w - \epsilon_\theta(x_t^w, t) \|_2^2 - \| \epsilon^w - \epsilon_{\theta_0}(x_t^w, t) \|_2^2 - \notag \\
    (\| \epsilon^l - \epsilon_\theta(x_t^l, t) \|_2^2 - \| \epsilon^l - \epsilon_{\theta_0}(x_t^l, t) \|_2^2) 
    ) ),
\end{align}

Here \(\epsilon_\theta\) is the denoising network, \(\epsilon_{\theta_0}\) the pretrained network, and $\omega(\lambda_t)$ absorbs all coefficients. This formulation improves \(\epsilon_\theta\) on denoising of preferred samples \(x_0^w\) more than dispreferred \(x_0^l\), aligning diffusion models with preferences without additional inference costs or mode collapse.

\subsection{DDIM Inversion}
\label{sec:DDIM_Inversion}
In this part, we introduce DDIM Inversion which is crucial in our proposed method. 
In traditional diffusion models like DDPM~\cite{ddpm}, the generative process starts with Gaussian noise $x_T$ and gradually denoises through a reverse Markov chain to obtain a clean image $x_0$. Although the training procedure maximizes a variational lower bound, the sampling process is inherently stochastic, making it difficult to precisely recover the sampling path $\{x_t\}$ for a given $x_0$.

Denoising Diffusion Implicit Models (DDIM)~\cite{song2020ddim} proposes a non-Markovian inference process. Given any intermediate state $x_t$, the trained $\epsilon_\theta(x_t, t)$ estimates the corresponding denoised sample:
\begin{equation}
    \hat{x}_0=f_\theta^{(t)}(x_t) = \frac{x_t - \sqrt{1-\alpha_t}\,\epsilon_\theta(x_t, t)}{\sqrt{\alpha_t}},
    \label{eq: x_0_predicton}
\end{equation}
where $\alpha_t$ is a signal retention coefficient associated with time step $t$.
When $t>1$, the generative process is given by
\begin{equation}
\label{eq:ddim_sampling}
\begin{aligned}
    &p_\theta(x_{t-1}|x_{t})=\mathcal{N}\left(\sqrt{\alpha_{t - 1}}\hat{x}_0+\sqrt{1 - \alpha_{t - 1}-\sigma_{t}^{2}}\cdot\frac{\boldsymbol{x}_{t}-\sqrt{\alpha_{t}}\hat{x}_0}{\sqrt{1 - \alpha_{t}}}, \sigma_{t}^{2}\boldsymbol{I}\right)\\
\end{aligned}
\end{equation}
When $t=1$, $p_\theta(x_{0}|x_{1}) =\mathcal{N}\left(\hat{x}_0, \sigma_1^2 \boldsymbol{I}\right)$. The noise term $\sigma_t$ can be zero during sampling, thus converting the generative process into a deterministic implicit probabilistic model. Since the update process no longer introduces additional randomness, DDIM not only allows high-quality samples to be generated in fewer steps but also renders the generative mapping invertible, which enables recovering the sampling trajectory $\{x_1, x_2, \ldots, x_T\}$ from $x_0$~\cite{song2020ddim}.

% Using this denoising estimation, the iteration under the deterministic setting ($\sigma_t = 0$) in DDIM is
% \begin{equation}
% \begin{aligned}
%     x_{t-1} &= \sqrt{\alpha_{t-1}} \, f_\theta^{(t)}(x_t) + \sqrt{1-\alpha_{t-1}}\,\epsilon_\theta(x_t, t).\\
%     &= \frac{\sqrt{\alpha_{t - 1}}}{\sqrt{\alpha_t}}\boldsymbol{x}_t+\left(\sqrt{1 - \alpha_{t - 1}}-\sqrt{\frac{\alpha_{t - 1}}{\alpha_t} -\alpha_{t - 1} }\right)\epsilon_{\theta}(\boldsymbol{x}_t, t)
% \end{aligned}
%     \label{eq: generative_formula}
% \end{equation}
% The mapping from $x_t$ to $x_{t-1}$ is unique. 
It has been observed that when the number of sampling steps is large, predicted noises are close in adjacent steps~\cite{sferd}. This supports the assumption that the ordinary differential equation (ODE) process can be reversed within the limit of small steps~\cite{DiffusionCLIP}, which means $\epsilon_\theta(x_t, t) \approx \epsilon_\theta(x_{t-1}, t-1)$. 
This key property allows us to invert the DDIM generative process~\cite{Prompt_Tuning_Inversion,directinversion}. Starting from a real sample $x_0$, we can backtrack to sampling its potential denoising sequence $\{x_1, x_2, \ldots, x_T\}$. 
% The core idea of the inversion process is as follows:
% \begin{enumerate}
%     \item Use the noise prediction network $\epsilon_\theta(\cdot, t)$ to compute the denoising estimate $f_\theta^{(t)}(x_t)$ for any $x_t$, thereby obtaining a “predicted” original image.
%     \item Exploit the invertibility of the update equation (1) to iteratively “backtrack” from $x_0$ to recover the latent states $x_t$.
% \end{enumerate}
% Thus, one can derive an implicit mapping from $x_0$ to any time step $t$.
The detailed inversion iteration is
% The inversion process can be formulated as
\begin{equation}
    x_t=\sqrt{\frac{\alpha_t}{\alpha_{t-1}}}x_{t-1}+(\sqrt{1-\alpha_t}-\sqrt{\frac{\alpha_t}{\alpha_{t-1}}-\alpha_t})\epsilon_\theta(x_{t-1},t-1)
    \label{eq: DDIM-inverison}
\end{equation}
Such iteration finally reaches $x_T$. Under the assumption above, with $x_T$ as the initial noise, a deterministic sampling process $\epsilon_\theta$ is likely to recover the sequence \(\{x_{T-1}, \ldots, x_1 \}\) and come back to the original $x_0$.
More advanced estimations~\cite{ReNoise,directinversion} give more precise recovery.
This capability to recover the latent path from $x_0$ provides a theoretical basis for Inversion-DPO we introduce next.

% **Key Innovation**: Both methods replace multi-stage RLHF with direct policy optimization. DPO reformulates reward-policy duality via preference likelihood, while Diffusion-DPO adapts this to diffusion models by integrating ELBO and trajectory-based KL constraints.

\section{Method}

\begin{figure*}[t]
    \centering
    \includegraphics[width=1\textwidth]{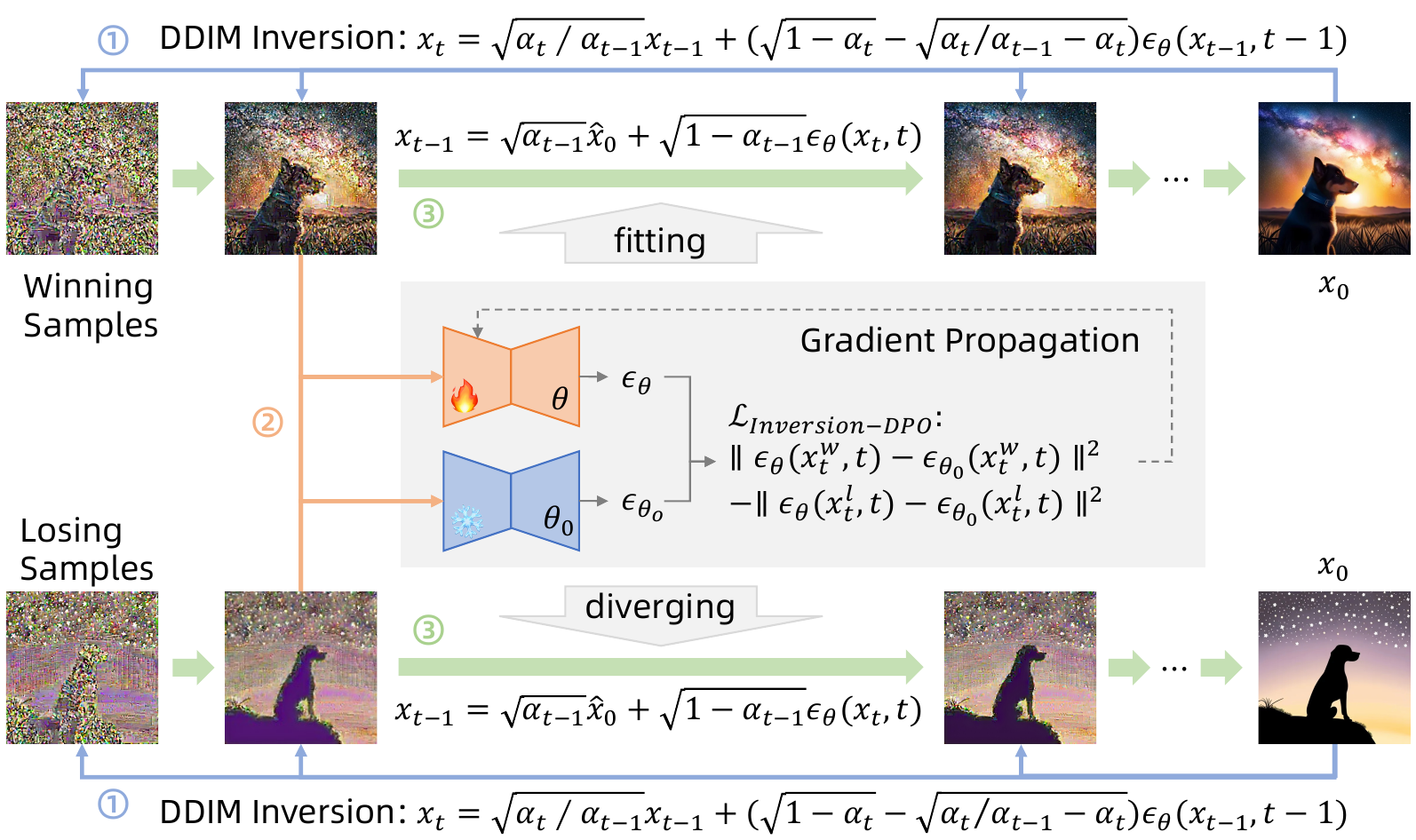} % Reduce the figure size so that it is slightly narrower than the column.
    % \vspace{-1.5em}
    \caption{Computation pipeline of the proposed Inversion-DPO. 1) Given an initial image \( x_0 \), we obtain the diffusion trajectories \( \{x_1, x_2, \ldots, x_T\} \) of both the winning and losing samples from a reference model via DDIM inversion. 2) At each timestep along the diffusion trajectory, the model is trained to predict the noise \( \epsilon \), aligning it closely with the \( \epsilon \) of the winning sample while diverging from that of the losing sample. 3) This process is repeated across the entire diffusion trajectory, continuously updating model parameters until the final predicted image is generated.}
    %计算流程 of 我们提出的Inversion-DPO。1）给定x_0，我们可以通过引入DDIM inversion得到winning和losing sample在扩散轨迹上的分布$\{x_1, x_2, \ldots, x_T\}$from reference model。2）对于扩散轨迹上每一步，训练模型预测epsilon，并令其拟合winning样本对应的epsilon，远离losing样本的epsilon。3）在整个扩散轨迹上重复上一步的计算，不断更新模型参数，直到得到最终的predicted图像。}
    \label{fig: pipeline}
    % \vspace{-1em}
\end{figure*}

\subsection{Inversion Direct Preference Optimization}
% loss of DPO
DPO~\cite{rafailov2023directDPO} is an effective reward-guided post-training approach that directly learns human preferences through a reparameterized policy, rather than optimizing a reward function and then performing reinforcement learning (RL). 
%解释参数

%当adapting DPO to diffusion models, 目标从原始的拟合x_0变为拟合路径（x_1,...,x_T）。目标变为：
Recall that when adapting DPO to diffusion models~\cite{Wallace2024DiffusionDPO} , the objective shifts from fitting \( x_0 \) to fitting the trajectory \((x_0, \dots, x_T)\)with 
% loss of diffusion-DPO
\begin{equation}
\label{eq:diffusion-dpo-2}
\begin{aligned}
 & \mathcal{L}_{\text{Diffusion-DPO}}(\theta)=-\mathbb{E}_{(\boldsymbol{x}_0^w,\boldsymbol{x}_0^l)}\log\sigma\bigg( \\
 & \beta\mathbb{E}_{\substack{ \boldsymbol{x}_{1:T}^w\sim p_\theta(\boldsymbol{x}_{1:T}^w|\boldsymbol{x}_0^w) \\ 
 \boldsymbol{x}_{1:T}^l\sim p_\theta(\boldsymbol{x}_{1:T}^l|\boldsymbol{x}_0^l)}}
 \left[\log\frac{p_\theta(\boldsymbol{x}_{0:T}^w)}{p_{\theta_0}(\boldsymbol{x}_{0:T}^w)}-\log\frac{p_\theta(\boldsymbol{x}_{0:T}^l)}{p_{\theta_0}(\boldsymbol{x}_{0:T}^l)}\right]\bigg)
\end{aligned}\end{equation}
%拟合这个目标，要已知扩散轨迹上所有时刻的数据分布。因此我们将DDIM inversion，一种预测扩散轨迹的方法，引入DPO-Diffusion，以简化及精确原始的计算。首先将
To optimize this, $\boldsymbol{x}_{1:T}$ across all time steps are required with a posterior sampling of $p_\theta$.
% to model the noisy distribution $p_\theta(\boldsymbol{x}_{1:T})$ across all time steps along the diffusion trajectory. 
Thus, Diffusion-DPO encounters two key limitations: 1) The intractability of $p_\theta(\boldsymbol{x}_{1:T}|\boldsymbol{x}_0)$ results in a subtitution with $q(\boldsymbol{x}_{1:T}|\boldsymbol{x}_0)$ as approximation, introducing large estimation errors; 2) The KL-Divergence formulation expands the original pairwise comparison into four terms. 

We address these issues by designing Inversion-DPO~(\cref{fig: pipeline}). Our key insight stems from the deterministic characteristic of DDIM Inversion~\cite{song2020ddim}. 
Given a sample $\boldsymbol{x}_0$, DDIM Inversion constructs the sampling trajectory $\boldsymbol{x}_{1:T}$ leading to $\boldsymbol{x}_0$ through a deterministic mapping by equation~\eqref{eq: DDIM-inverison}. 
Thus, we propose to approximate the posterior $p_\theta(\boldsymbol{x}_{1:T}|\boldsymbol{x}_0)$ with DDIM inversion.
This eliminates the need for stochastic approximation $q$ in trajectory estimation.
Using DDIM Inversion is reasonable with three advantages.
First, assuming the deterministic sampling setting, the posterior inference is the inversion process. Thus, the resulting $\boldsymbol{x}_{1:T}$ are more precise.
Second, such inversion is compatible with the definition of the reward function on diffusion in equation~(\ref{eq:reward_maximization}), so the implicit reward learning is also precise.
Third, the final estimated expectation in~(\ref{eq:diffusion-dpo-2}) is also more precise in this case.

Given the $\boldsymbol{x}_{1:T}^w$, we derive the following computation within the expectation.
Specifically, $\log\frac{p_\theta(\boldsymbol{x}_{0:T}^w)}{p_{\theta_0}(\boldsymbol{x}_{0:T}^w)}$ is first decomposed into the differences in conditional probabilities at each time step:
\begin{equation}
\log\frac{p_\theta(\boldsymbol{x}_{0:T}^w)}{p_{\theta_0}(\boldsymbol{x}_{0:T}^w)}
=
\sum_{t=1}^T\left[\log p_\theta(\boldsymbol{x}_{t-1}^w|\boldsymbol{x}_t^w)-\log p_{\theta_0}(\boldsymbol{x}_{t-1}^w|\boldsymbol{x}_t^w)\right]
\end{equation}
% 因x和y的a分布相同，故抵消。
Since \( p_{\theta} \) and \( p_{\theta_0} \) share the same distribution of \( x_T \) as standard Gaussian, $\log p_{\theta}(x_T^w) - \log p_{\theta_0}(x_T^w)$ cancel out. 
% 使用$p_\theta$的DDIM inversion之后得到的 \((x_1, \dots, x_T)\)，其中给定$x_t$，根据公式（7）和（9），每一个采样出来的$x_{t-1}$都是确定性的，可以写成：
% After applying DDIM inversion under \( p_\theta \), we obtain the sequence \( (x_1^w, \dots, x_T^w) \).
The conditional probabilities of \( p_{\theta} \) and \( p_{\theta_0} \) are Gaussian distributions as in equation~(\ref{eq:ddim_sampling}) and becomes dirac-delta as $\sigma_t$ approaches $0$.
We first continue derivation with $\sigma_t\neq 0$ and adopt a simple approximation~\cite{ddpm} of the final loss.

Given the assumption of inversion in Sec.~\ref{sec:DDIM_Inversion}, the values of $\boldsymbol{x}_{t-1}^w$ are similar in the inversion from $x_0^w$ and in the sampling from $x_T^w$. 
Therefore, $\log p_\theta(\boldsymbol{x}_{t-1}^w|\boldsymbol{x}_t^w)$ is only related to $\sigma_t^2$ and thus constant. 
At the same time, given 
\begin{align}
    &\boldsymbol{\mu}_{\theta_0}(x_t^w) = \frac{\sqrt{\alpha_{t - 1}}}{\sqrt{\alpha_t}}\boldsymbol{x}_t^w + \notag \\
    &\left(\sqrt{1 - \alpha_{t - 1} - \sigma_t^2}-\frac{\sqrt{\alpha_{t - 1}({1 - \alpha_t})}}{\sqrt{\alpha_t}}\right)\epsilon_{\theta_0}(\boldsymbol{x}_t^w, t)
\end{align}
we have the following (details included in supplementary material):
\begin{equation}
    \begin{aligned}
        -\log p_{\theta_0}(\boldsymbol{x}_{t-1}^w|\boldsymbol{x}_t^w) &\propto  \|\boldsymbol{\mu}_{\theta_0}(x_t^w)-\boldsymbol{x}_{t-1}^w\|^2 \\
        &\propto 
        \|\epsilon_{\theta_0}(\boldsymbol{x}_t^w,t)-\epsilon_{\theta}(\boldsymbol{x}_t^w,t)\|^2\\
    \end{aligned}
\end{equation}
Summing the contributions across all time steps, we can arrive at:
\begin{equation}
\log\frac{p_\theta(\boldsymbol{x}_{0:T}^w)}{p_{\theta_0}(\boldsymbol{x}_{0:T}^w)} \propto
\left[\sum_{t=1}^T\|\epsilon_\theta(\boldsymbol{x}_t^w,t)-\epsilon_{\theta_0}(\boldsymbol{x}_t^w,t)\|^2\right], \end{equation}

The term for the losing sample in equation (\ref{eq:diffusion-dpo-2}) is derived analogously. 
Each term in the objective can be computed exactly. The log-probability ratio now directly captures the policy divergence without introducing approximation errors.
% it is obvious that 目标中的每一项都可以被精确计算得出，The log-probability ratio now directly reflects policy divergence without approximation errors。

Taking all together, the final objective becomes:
\begin{equation}\begin{aligned}
 \mathcal{L}_{\text{Inversion-DPO}}(\theta)=-\mathbb{E}_{(\boldsymbol{x}_0^w,\boldsymbol{x}_0^l)}\log\sigma\bigg(  \beta\mathbb{E}_{\substack{ \boldsymbol{x}_{1:T}^w\sim p_\theta(\boldsymbol{x}_{1:T}^w|\boldsymbol{x}_0^w) \\ 
 \boldsymbol{x}_{1:T}^l\sim p_\theta(\boldsymbol{x}_{1:T}^l|\boldsymbol{x}_0^l)}}\\
\sum_{t=1}^T\Bigg[ \|\epsilon_\theta(\boldsymbol{x}_t^w,t)-\epsilon_{\theta_0}(\boldsymbol{x}_t^w,t)\|^2- \|\epsilon_\theta(\boldsymbol{x}_t^l,t)-\epsilon_{\theta_0}(\boldsymbol{x}_t^l,t)\|^2\Bigg]\bigg)
\end{aligned}\end{equation}
% where $f(\boldsymbol{x}_{1:T}^w) = \mathbb{E}_{\boldsymbol{x}_0^w}\left[\sum_{t=1}^T\frac{1-\alpha_t}{2\sigma_t^2\alpha_t}\|\epsilon_\theta(\boldsymbol{x}_t^w,t)-\epsilon_{\theta_0}(\boldsymbol{x}_t^w,t)\|^2\right]$, and the same applies to $f(\boldsymbol{x}_{1:T}^l)$.
% where $\omega_t = \frac{1-\alpha_t}{2\sigma_t^2\alpha_t}$ are precomputed weights. 

Notably, during the training of Inversion-DPO, the previous four-term loss collapses into two deterministic summations due to trajectory determinism, which is conceptually simpler. This simplification also yields non-negligible improvements in optimization efficiency for model training. 
The pretrained $\epsilon_{\theta_0}$ can be transformed to approximate the score.
Fitting the noise added as in (\ref{eq:diffusion-DPO-final}) introduces extra variance~\cite{6795935,Diffusion_based_representation_learning} which in unstable in the training~\cite{STF}, while fitting the predicted noise close to the score is more stable and thus improves the training efficiency.
Our training schema fundamentally resolves the precision-efficiency tradeoff in preference-based diffusion alignment.

\subsection{Inversion-DPO with Multiple Objectives}
\label{sec:dataset_collection}
% For the learning of Inversion-DPO, both positive (i.e., winning samples) and negative (i.e., losing samples) data pairs are required to guide the optimization process. However, in task-specific scenarios, the evaluation of samples often involves complex conditions where a single metric may not provide an effective assessment. 
For effective Inversion-DPO training, both positive (i.e., winning samples) and negative (i.e., losing samples) data pairs are required to guide the optimization. While manual annotation is commonly used to obtain such pairs~\cite{zhang2024itercomp,Wallace2024DiffusionDPO}, it is costly and prone to ambiguity. To address this, recent works~\cite{tian2025diffusionSharpening,Black2024DDPO,Yang2024D3PO,ma2025inferencetimescalingdiffusionmodels} adopt automatic metrics to rank images as guidance. 
We follow this strategy.
However, relying on a single metric is often inadequate to comprehensively capture the quality of the outputs, especially in complex scene generation scenarios. Therefore, we introduce a multi-objective learning strategy, leveraging multiple metrics to provide more reliable and informative preference signals.

% Specifically, we incorporate the DPG-Bench Score and the 3-in-1 score from T2I-CompBench, which serve as representative evaluation metrics in complex scenarios, to compute the reward scores for data pairs. The corresponding reward score calculation is as follows:
Specifically, we incorporate multiple evaluation metrics to derive the reward scores for data pairs, computed as follows:
\begin{equation}
    r(c,x_0)=\frac{1}{N}\sum_{i=1}^{N}r_i\left(c,x_0\right)
\end{equation}
Here, \( N \) denotes the number of evaluation metrics utilized in the current task, and \( (c, x_0) \) represents the conditional sample pair generated by \( p_{\theta_0} \) given the condition \( c \).

% 直接衔接数据集构建部分

We evaluate the effectiveness of Inversion-DPO in two representative tasks. In addition to the typical text-to-image generation, we further assess it in the more challenging setting of compositional image generation, which requires the model to accurately synthesize complex scenes involving multiple objects and relationships. Existing studies~\cite{Li2024LAIONSG,wang2024scene} suggest that, compared to textual annotations, structured annotations provide a more compact and explicit form of information integration, making them more effective for representing complex scenes. Such structured annotations are typically represented as graphs, consisting of nodes and edges that denote objects in the scene and their interactions, respectively.  

% 现有的Pick-a-Pic数据集可以支持Inversion-DPO在basic的t2i任务上的训练，然而对于compositional generation，没有一个对应的带有pair的数据集可以采用，since existing datasets with structured annotations~\cite{caesar2018coco,krishna2017visual,Li2024LAIONSG} only provide single-condition samples. 所以we construct a compositional paired dataset with structured annotations to support model training in this domain.  

On basic text-to-image tasks, Pick-a-Pic dataset~\cite{Yuval2023Pick-a-Pic} serves as a reliable source of supervision for training Inversion-DPO. However, for compositional generation, there is a lack of corresponding paired datasets that can be directly utilized. This is because existing datasets with structured annotations~\cite{caesar2018coco,krishna2017visual,Li2024LAIONSG} provide only single-condition samples, making them insufficient for preference-based learning. To bridge this gap, we construct a paired dataset with structured annotations to support alignment and evaluation.

In detail, for a given complex scene annotation, we generate four samples using an existing baseline~\cite{Li2024LAIONSG}, and compute a multi-dimensional score for each generated image. During training, the model determines positive and negative samples based on a combination of these scores, guiding the learning of Inversion-DPO.  

To ensure effective evaluation of complex scene images, we adopt two representative metrics--DPG-Bench Score~\cite{hu2024ella} and the 3-in-1 score from T2I-CompBench~\cite{T2I_CompBench}--as the criteria for multi-objective learning. In total, we generate 11,140 images for 2,785 annotations, with each image associated with a comprehensive complex scene score. Rather than enforcing fixed pairwise bindings, we allow the model to dynamically determine relative ``winning'' and ``losing'' samples based on their comprehensive scores during training. This approach prevents the dataset size from being constrained to the original 2,785 samples, enabling scalable data expansion within a certain range and ultimately improving training quality.

\section{Experiments}

\begin{figure}[t]
    \centering
    \includegraphics[width=0.9\columnwidth]{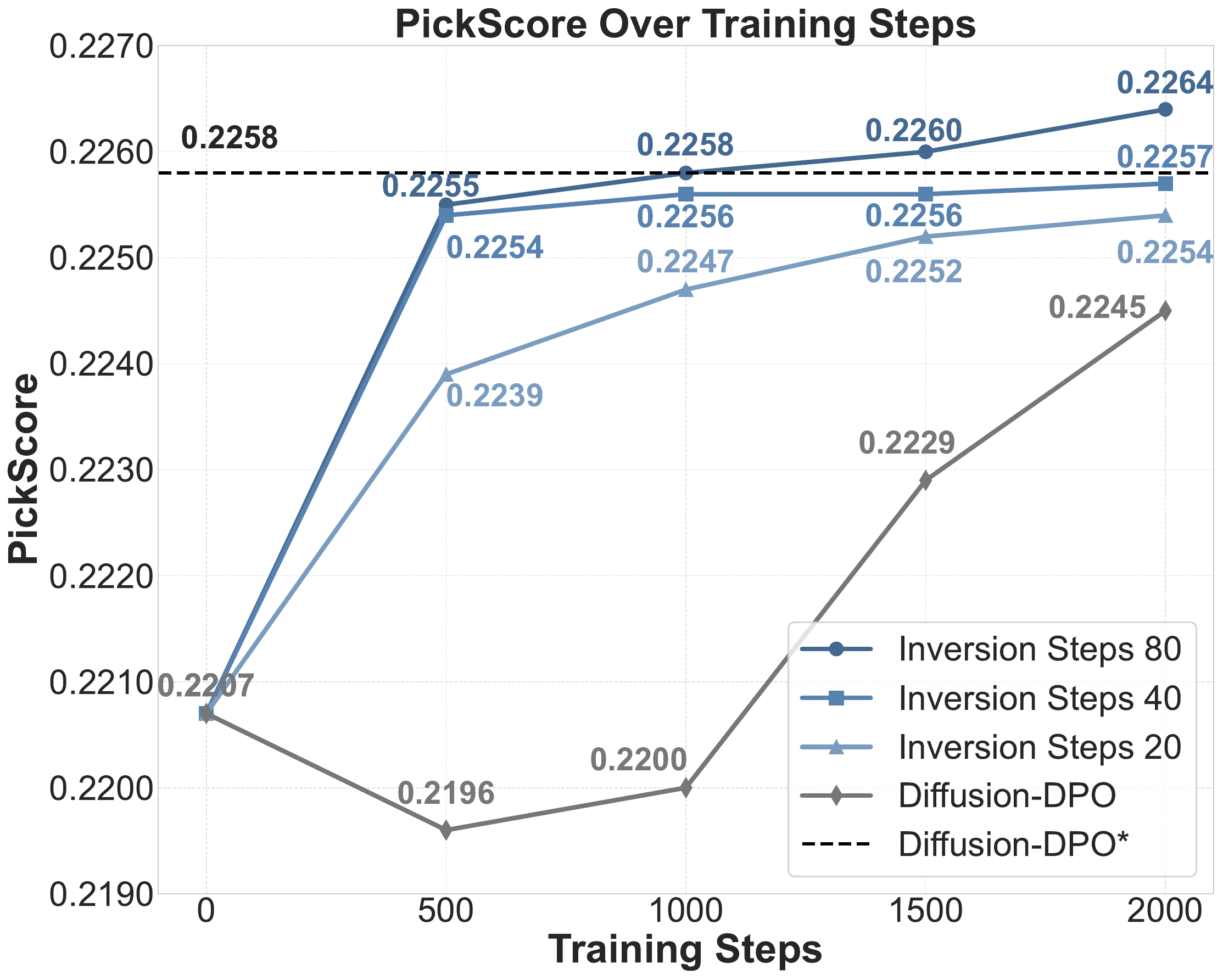} % Reduce the figure size so that it is slightly narrower than the column.
    % \vspace{-1.5em}
    \caption{A demonstration of how PickScore increase with more training steps. The blue curve shows the performance of our Inversion-DPO trained with different numbers of DDIM inversion steps. The gray curve represents our retraining with the official implementation of Diffusion-DPO~\cite{Wallace2024DiffusionDPO}, while the dashed line corresponds to the officially released weights. Our model with 80-step inversion outperforms the both the retraining and the released Diffusion-DPO checkpoint. All models are trained under identical configurations.}
    % 模型的训练steps与PickScore增长程度关系图。蓝色折线是我们的Inversion-DPO采用不同DDIM inversion步数训练的结果。 灰色折线代表我们训练的Diffusion-DPO， 虚线代表官方发布权重的Diffusion-DPO。我们的实验中所有模型采用相同的训练配置。}
    \label{fig: efficiency}
    % \vspace{-1em}
\end{figure}

\begin{figure}[t]
    \centering
    \includegraphics[width=0.9\columnwidth]{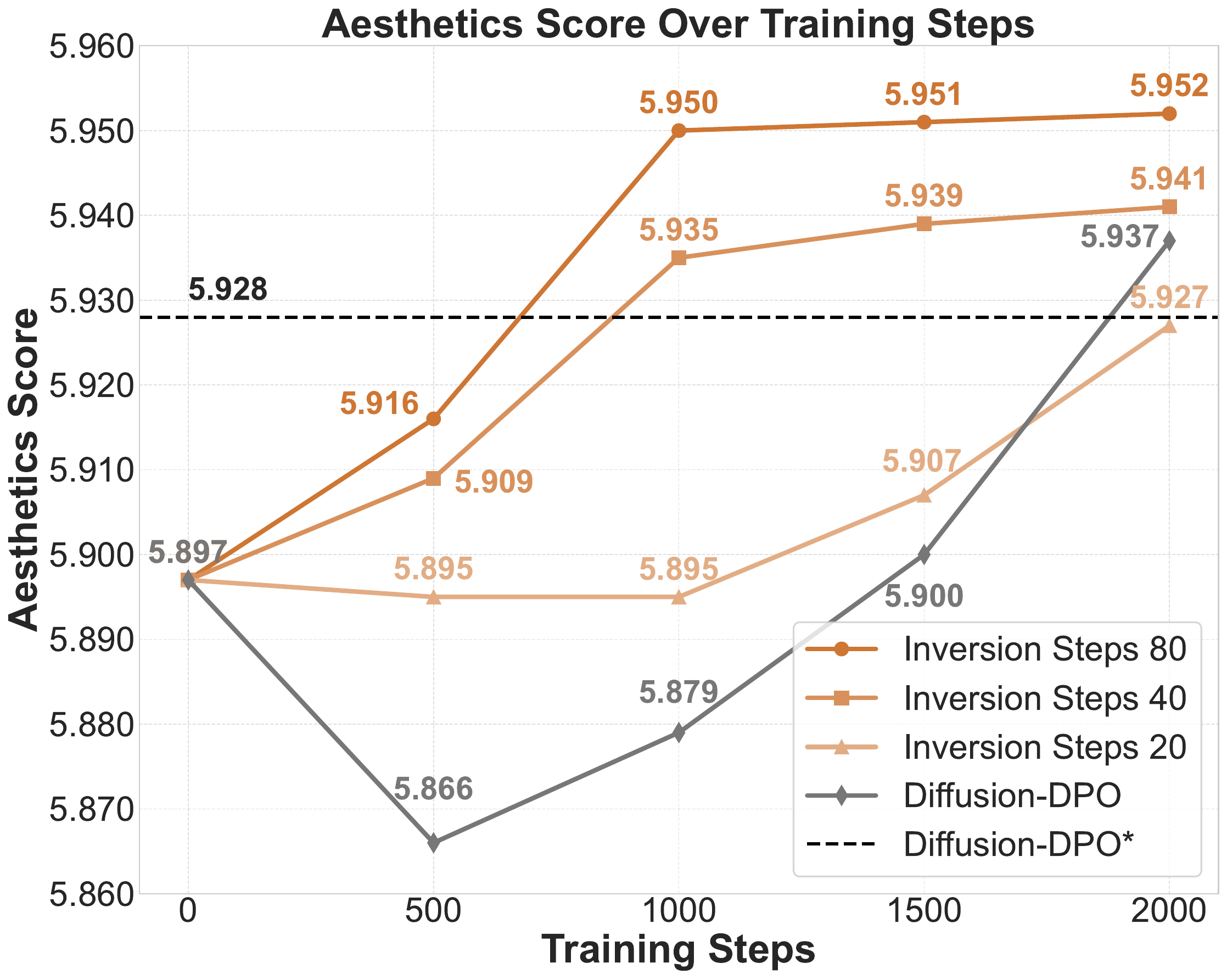} % Reduce the figure size so that it is slightly narrower than the column.
    \caption{Relationship between training steps and Aesthetics Score improvements. The orange curve shows the performance of our Inversion-DPO trained with different numbers of DDIM inversion steps. The gray curve represents our implementation of Diffusion-DPO~\cite{Wallace2024DiffusionDPO}, while the dashed line corresponds to the officially released weights of Diffusion-DPO. All models are trained under identical configurations.}
    \label{fig: efficiency_as_score}
\end{figure}

\subsection{Dataset and Implementation Details}
%T2I任务是pick-a-pic数据集，sg2im是LAIN-SG数据集中采样的complex部分。
%Inversion-DPO在上述两个数据集上进行训练。实现细节，基于SDXL，XXXXX

%dataset：我们使用Pick-aPic v2数据集训练T2I模型，该数据集包含SDXL-beta和Dreamlike生成的图像对偏好以及从网络收集的提示和偏好对。采用与Diffusion-DPO相同的数据处理方式，排除12%的平局样本，最终获得851,293个有效数据对。对于SG2IM任务，我们从LAION-SG中精选图像对进行训练，确保数据质量和多样性(xx章xx节）。

%training：训练过程中，我们为SDXL和SGXL-SG采用Adafactor优化器以节省内存，而SD1.5训练则使用AdamW。所有模型的学习率统一设为1e-6。特别地，Inversion-DPO中的β参数被设置为2000，Inversion Step被分别设置为20，40，80，以观察不同的DDIM Inversion步数对模型性能的影响（xx章xx节）。每个模型均训练2000步，在2张H800 GPU上完成所有训练任务。

\textbf{Dataset.} We use the Pick-a-Pic v2 dataset~\cite{Yuval2023Pick-a-Pic} to train our T2I model, which contains paired preferences of images generated by SDXL-beta and Dreamlike, along with prompts and preference pairs collected from the web. Following the same data processing approach as Diffusion-DPO~\cite{Wallace2024DiffusionDPO}, we exclud 12\% of tie samples, ultimately obtaining 851,293 effective data pairs. For the compositional image generation task, as described in Section~\ref{sec:dataset_collection}, we curate compositional image pairs from LAION-SG~\cite{Li2024LAIONSG} for training, ensuring data quality and diversity.

\textbf{Training.} During the training process, we employ the Adafactor~\cite{shazeer2018adafactoradaptivelearningrates} optimizer for SDXL and SGXL-SG to save memory, while AdamW~\cite{loshchilov2019decoupledweightdecayregularization} was used for SD1.5 training. The learning rate for all models was uniformly set to 1e-6. Specifically, the $ \beta$
parameter in Inversion-DPO was set to 2000. We also set the Inversion Steps to 20, 40, and 80, respectively, to observe the impact of different DDIM Inversion steps on model performance, which is detailed in our ablation study. Each model was trained for 2000 steps, with all training tasks completed on two NVIDIA H800 GPUs.

\subsection{Baselines and Evaluation Metrics}
% 对于传统的从文本到图像生成，我们选择典型的text-to-image模型以及一些后训练方法，包括SD1.5, SDXL, Diffusion-DPO, DDPO, D3PO, Demon和IterComp. 我们用PickScore, CLIP Score以及Aesthetic Score对生成结果进行综合评估，分别衡量图像质量与人类偏好对齐的程度、与输入prompt的语义一致性、以及图像的细节美学质量。
For the typical text-to-image generation task, we evaluate a set of representative baseline models and post-training methods, including SD1.5~\cite{ldm}, SDXL~\cite{podell2023sdxlimprovinglatentdiffusion}, Diffusion-DPO~\cite{Wallace2024DiffusionDPO}, DDPO~\cite{Black2024DDPO}, D3PO~\cite{Yang2024D3PO}, Demon~\cite{yeh2025Demon}, and IterComp~\cite{zhang2024itercomp}. The generated results are assessed using PickScore~\cite{Yuval2023Pick-a-Pic}, CLIP Score~\cite{Radford2021Learning_CLIP}, and Aesthetic Score~\cite{LAION_5B}, which respectively measure alignment with human preference, semantic consistency with the input prompt, and the aesthetic quality of image details.

% 对于compositional image generation, 我们评估了SGDiff, SG-Adapter, SDXL-SG以及SD1.5-SG。考虑到复杂场景生成的目标，我们采用的评估准则有图像质量指标FID，以及分别衡量图像中场景一致性、object一致性和object间关系一致性的SG-IoU, Entity-IoU and Relation-IoU.

For compositional image generation, we evaluate SGDiff~\cite{Yang2022DiffusionBasedSG}, SG-Adapter~\cite{Shen2024SGAdapterET}, SDXL-SG~\cite{Li2024LAIONSG}, and SD1.5-SG~\cite{Li2024LAIONSG}. Considering the objectives of complex scene generation, we adopt FID~\cite{FID} as the image quality metric, along with SG-IoU, Entity-IoU, and Relation-IoU~\cite{Shen2024SGAdapterET}, which respectively assess scene-level consistency, object-level consistency, and the consistency of relationships between objects within the generated images.

\subsection{Qualitative Results}

\begin{figure*}[t]
    \centering
    \includegraphics[width=1\textwidth]{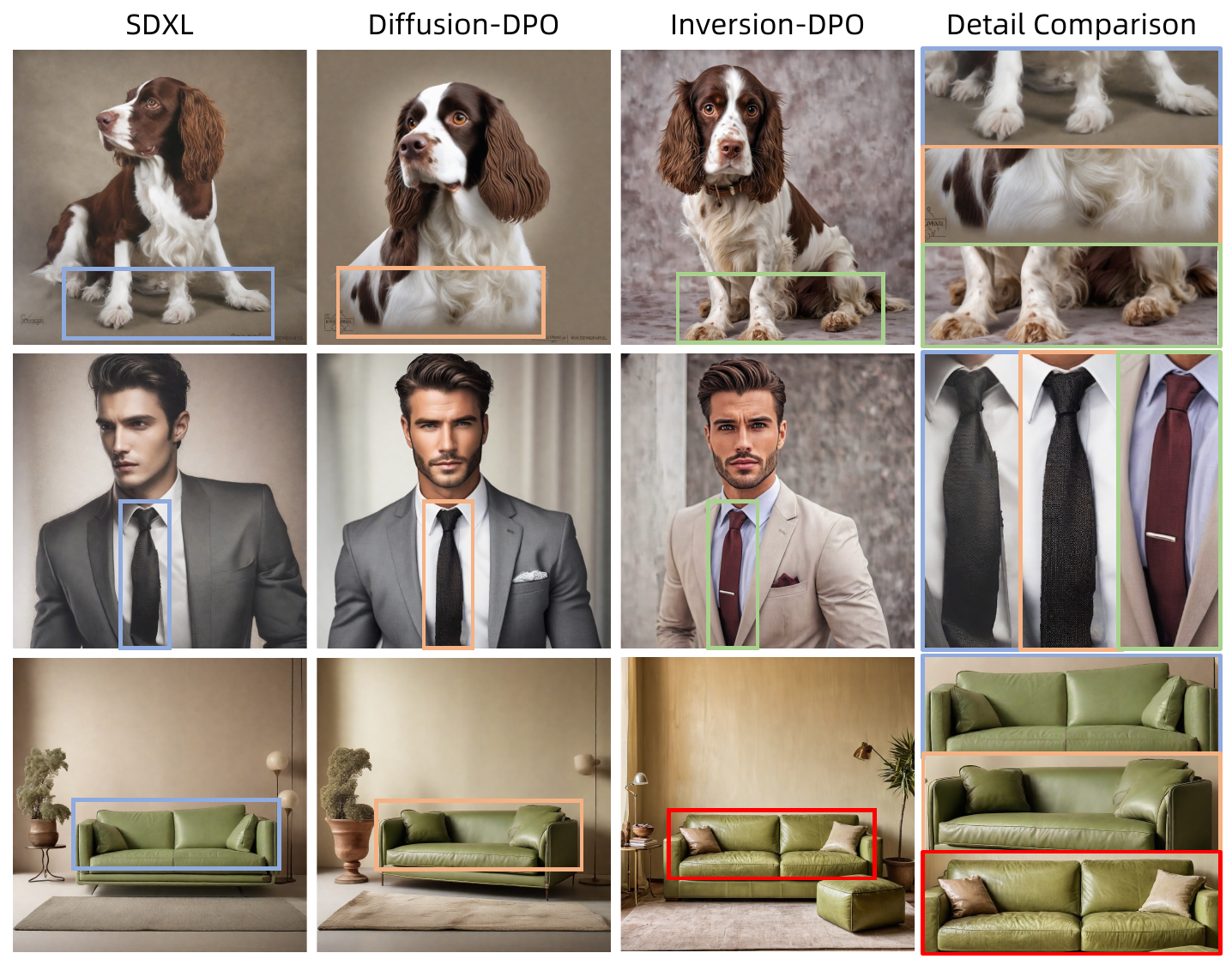} % Reduce the figure size so that it is slightly narrower than the column.
    % \vspace{-1.5em}
    \caption{A comparison of generation details. Existing models often struggle to produce fine-grained details, while Inversion-DPO demonstrates a higher level of fidelity in generating nuanced elements such as limbs, ties, and textures.}
    \label{fig: qualitative-detail}
    % \vspace{-1em}
\end{figure*}

%\begin{figure*}[t]
%    \centering
%    \includegraphics[width=1\textwidth]{acmart-primary/samples/figures/sg2im-qualitive.pdf} % Reduce the figure size so that it is slightly narrower than the column.
%    % \vspace{-1.5em}
%    \caption{Qualitative results on compositional image generation. Inversion-DPO proves effective in addressing key challenges in compositional image generation, including object omission (orange and green annotations in (a), (d), and (e)), attribute errors (red annotations in (b), (c), and (d)), and relation inaccuracies (blue annotations in (c) and (d)). As shown in the last row, Inversion-DPO achieves satisfactory generation results on cases where baseline models struggle.}
%    % compositional image generation的quallitative result。Inversion-DPO在解决compositional image generation的现存问题上是有效的。这些问题包括：object缺失（（a）（d）（e）中的橙色和绿色标注）,属性错误（（b）(c)(d)中的红色标注），和关系错误（(c)（d）中的蓝色标注）。 baseline难以handle的问题Inversion-DPO都做到了令人满意的生成（最后一行）。}
%    \label{fig: qualitative-sg2im}
%    % \vspace{-1em}
%\end{figure*}

% t2i, sg2im, detailed example, waiting for images.
%fig X 展示了我们的Inversion-DPO和其他基线模型在基础的text-to-image任务上的视觉表现。每一行是输入的prompt和对应的生成图像。相比于现有baselines, Inversion-DPO具有极高的视觉吸引力，并且更容易达到最高的PickScore。此外，我们提出的方法在细节生成的把控which是一个t2i模型难以妥善handle的困境上展示出promising潜力，如fig Y所示。

\Cref{fig: teaser} illustrates the visual performance of our Inversion-DPO compared to baseline models on text-to-image tasks. Each row shows an input prompt and the corresponding generated images. Compared to existing baselines~\cite{podell2023sdxlimprovinglatentdiffusion,Black2024DDPO,yeh2025Demon,Wallace2024DiffusionDPO}, Inversion-DPO exhibits higher visual appeal and achieves higher PickScore. Moreover, in \cref{fig: qualitative-detail}, our method demonstrates promising potential in controlling fine-grained detail generation, which is a persistent challenge for text-to-image models. We further visualize results on more challenging compositional image generation tasks. 
%As shown in \cref{fig: qualitative-sg2im}, Inversion-DPO demonstrates superior performance in terms of object occurrence, relation accuracy, and attribute binding.
Moreover, Inversion-DPO demonstrates superior performance in terms of object occurrence, relation accuracy, and attribute binding (details included in supplementary material).
% 同时，我们可视化了在更加chanllenging的compositional image generation上的结果，fig X表明Inversion-DPO object occurrence，relation正确性和attribute 绑定方面都更加优越。

We also conduct a user study, and Inversion-DPO enjoys 66.53\% human preference compared to SDXL and 72.67\% versus Diffusion-DPO. More details are in supplementary material.
% 为了验证与人类认知的一致性，我们进行了一项用户实验，结果显示，与SDXL相比，我们得到了66.53\%的用户偏好，与Diffusion-DPO相比，我们得到了72.67\%的用户偏好。用户实验的细节请参考supplementary material。
\subsection{Quantitative Results}

\begin{table}
    \centering
    \begin{tabular}{@{}lccc@{}}
    \toprule
        \textbf{Method} & \textbf{PickScore}$^\uparrow$ & \textbf{CLIP}$^\uparrow$ & \textbf{Aesthetic$^\uparrow$} \\
        \midrule
        SDXL~\cite{podell2023sdxlimprovinglatentdiffusion} & 0.223 & 0.334 & 6.07 \\ 
        Diffusion-DPO~\cite{Wallace2024DiffusionDPO} & 0.227 & 0.340 & 6.05 \\ 
        DDPO~\cite{Black2024DDPO} & 0.222 & 0.336 & 6.10 \\ 
        D3PO~\cite{Yang2024D3PO} & 0.223 & 0.338 & 5.99 \\
        Demon~\cite{yeh2025Demon} & 0.220 & 0.336 & 6.15 \\
        IterComp~\cite{zhang2024itercomp} & \textbf{0.232} & 0.340 & 6.22 \\
        \textbf{Inversion-DPO} & \textbf{0.232} & \textbf{0.341} & \textbf{6.24} \\
        \midrule
        SD1.5~\cite{ldm} & 0.207 & 0.320 & 5.61 \\
        Diffusion-DPO$\mathcal{y}$ & \textbf{0.212} & 0.324 & 5.73 \\
        \textbf{Inversion-DPO}$\mathcal{y}$ & \textbf{0.212} & \textbf{0.325} & \textbf{5.74} \\
        \bottomrule
    \end{tabular}
    % \vspace{-0.5em}
    \caption{Quantitative results of typical text-to-image generation. $\mathcal{y}$ indicates that the model is built upon SD1.5. The same applies to the following tables.}
    \label{tab: quantitative-t2i}
    % \vspace{-1em}
\end{table}

\begin{table}
    \centering
    \begin{tabular}{@{}lcccc@{}}
    \toprule
        \textbf{Method} & \textbf{FID}$^\downarrow$ & \textbf{SG-IoU}$^\uparrow$ & \textbf{Ent.-IoU$^\uparrow$}  & \textbf{Rel.-IoU$^\uparrow$} \\
        \midrule
        SGDiff~\cite{Yang2022DiffusionBasedSG} & 35.8 & 0.304 & 0.787 & 0.698 \\ 
        SG-Adapter~\cite{Shen2024SGAdapterET} & 27.8 & 0.314 & 0.771 & 0.693 \\ 
        R3CD~\cite{Liu2024R3CD} & 27.0 & 0.342 & 0.803 & 0.715 \\ 
        SDXL-SG~\cite{Li2024LAIONSG} & 26.7 & 0.340 & 0.792 & 0.703 \\ 
        \textbf{Inversion-DPO$\ast$} & \textbf{25.1} & \textbf{0.354} & \textbf{0.830} & \textbf{0.739} \\ 
        \midrule
        SD1.5-SG & 56.3 & 0.179 & 0.614 & 0.530 \\
        \textbf{Inversion-DPO}$\mathcal{y}$$\ast$ & \textbf{54.9} & \textbf{0.184} & \textbf{0.624} & \textbf{0.565} \\
        \bottomrule
    \end{tabular}
    % \vspace{-0.5em}
    \caption{Quantitative results on the more challenging task of compositional image generation. $\ast$ means the results of Inversion-DPO trained on the compositional paired dataset.}
    \label{tab: quantitative-sg2im}
    % \vspace{-1em}
\end{table}

%model efficiency

% 对于经典的t2i任务上的表现
% 对于复杂场景图像生成任务上的表现
% 训练步骤与奖励分数之间的关系

% tab.X reports the quantitative performance on typical text-to-image generation of the baseline models and ours. 不论是相比于基础的t2i模型、t2i后训练方法还是training-free的方法，我们的Inversion-DPO在人类偏好（PickScore）、与prompt一致性（CLIP Score）和美学细节（Aesthetic）上都达到了state-of-art. 除了基于SDXL的Inversion-DPO formal版本，我们还在SD1.5上进行了适配性验证。结果表明我们提出的方法对于多种Diffusion Model都是通用的，并且效果提升明显。
\Cref{tab: quantitative-t2i} reports the quantitative performance on typical text-to-image generation for both baseline models and ours. Compared to base T2I models~\cite{podell2023sdxlimprovinglatentdiffusion,ldm}, post-training methods~\cite{Wallace2024DiffusionDPO,Black2024DDPO,Yang2024D3PO,zhang2024itercomp}, and training-free approach~\cite{yeh2025Demon}, Inversion-DPO achieves state-of-the-art results in terms of human preference alignment (PickScore), text-image consistency (CLIP Score), and aesthetic quality (Aesthetic Score). In addition to the formal SDXL-based version of Inversion-DPO, we also validate its adaptability on SD1.5. 

% Results demonstrate that our method is broadly applicable across different diffusion models and yields significant performance improvements.

% 除了基础的图像t2i 任务，我们还探索了提出的方法在更加challenging的compositional image generation上的表现。如tab.X 所示，用Inversion-DPO训练过后的复杂场景生成模型获得了最好的图像质量（FID），并且在复杂度评价指标，which在scene级别、entity级别以及relation级别上综合衡量生成图像的复杂性忠实度，上达到最好表现。

We further investigate the performance of the proposed method on compositional image generation. As in \cref{tab: quantitative-sg2im}, after fine-tuning on the compositional dataset, Inversion-DPO achieves the best image quality as measured by FID, outperforming existing baseline models. Moreover, it obtains the highest scores on complexity evaluation metrics, which comprehensively assess the fidelity of the generated images at the scene, entity, and relation levels.

% The above results demonstrate that Inversion-DPO is a general and fundamental post-training method for diffusion models, adaptable to various backbone architectures and capable of delivering strong performance on both basic and more complex generation tasks. 
Empirically, our method also brings notable improvements in training efficiency. In \cref{fig: efficiency} and \cref{fig: efficiency_as_score}, Inversion-DPO achieves superior performance with significantly fewer training steps compared to Diffusion-DPO. 
Due to hardware limitation, our reimplementation of Diffusion-DPO with the official implementation does match the performance of the official checkpoint. 
Besides, we observe a slight decay of performance at the begining which may be a result of the instability of Diffusion-DPO's training.
Nevertheless, under the same resources, our Inversion-DPO constantly improves the performance of the generative model.
Specifically, our variants with only 20-step DDIM Inversion reaches a performance of 0.2247 at 1000 steps, comparable to Diffusion-DPO's 0.2245 at 2000, over 2$\times$ faster.
Inversion-DPO with 80 steps is over 4$\times$ faster.
Furthermore, at 2000 steps Inversion-DPO with 80-step inversion also outperforms the reported performance of Diffusion-DPO as labeled by the dotted line, further demonstrating the advantages of our method in both precision and efficiency. Results on other metrics are in Supplementary Material.
 %以上结果表明Inversion-DPO是一个通用的fundamental的Diffusion Model后训练方法，适配多个基础模型，并且在基础任务甚至更加复杂的任务上都有良好表现。
 % Additionally，我们的方法不仅有益于模型训练的精确度，在模型训练的效率上也有不可忽视的提升。
 % 如图X所示，和baseline model相比，Inversion-DPO在较少训练步数的情况下就可以达到较高级的表现。由于硬件限制等原因，我们自己训练的Diffusion-DPO无法达到和官方发布一样好的性能。然而在有限的训练资源的情况下，我们的Inversion-DPO结果仍然超过了官方发布的Diffusion-DPO，进一步说明了我们的方法在精度和效率上的both贡献。

\subsection{Ablation Study}
\label{subsec:ab_study}

\begin{table}
    \centering
    \begin{tabular}{@{}cccc@{}}
    \toprule
        \# \textbf{Inversion Steps} & \textbf{PickScore}$^\uparrow$ & \textbf{CLIP}$^\uparrow$ & \textbf{Aesthetic}$^\uparrow$ \\
        \midrule
        20 & 0.228 & 0.340 & 6.08  \\ 
        40 & 0.229 & 0.341 & 6.13  \\ 
        80 & \textbf{0.232} & \textbf{0.341} & \textbf{6.24}  \\ 
        \bottomrule
    \end{tabular}
    % \vspace{-0.5em}
    \caption{Effect of inversion steps on model performance.}
    \label{tab: ablation-inversion-steps}
    \vspace{-10pt}
\end{table}

% pq差异的影响
% inversion步数的影响

% Diffuion-DPO中, 由于逆向路径的不可访问性，approximate the reverse process $p_θ(x_{1:T}|x_0)$ with the forward $q(x_{1:T}|x_0)$。我们认为这一approximation是会引入不可忽视的训练误差的，为此，消融试验is conducted来验证这一影响。由于Inversion-DPO是多个time steps 反演噪声路径，而原始的Diffusion-DPO中是在$q(x_{1:T}|x_0)$中随机采样一个时间步骤的作为最终结果。为了比较的公平性，我们在Inversion-DPO中也随机采样$p_θ(x_{1:T}|x_0)$中的一个时间状态进行评估。如表X中present的，XXXXX

\textbf{Analysis of the different inversion steps.} We analyze the impact of different DDIM inversion steps on model training, with results shown in \cref{tab: ablation-inversion-steps} and \cref{fig: efficiency}. It is evident that as the number of inversion steps increases, the model consistently achieves better performance across all metrics. This can be attributed to the fact that with more DDIM inversion steps, the distributional gap between adjacent steps becomes smaller, satisfying the assumption in Sec.~\ref{sec:DDIM_Inversion} and making it easier to guide the model toward correct optimization directions. The configuration with 80 inversion steps is ultimately selected for training Inversion-DPO for a balance between performance and computational cost.

\section{Conclusion}

We presents Inversion-DPO, a novel post-training framework that integrates DDIM Inversion with DPO to achieve efficient and precise alignment in diffusion models. By leveraging the deterministic trajectory recovery of DDIM inversion, our method reduces approximation errors, achieving more than 2$\times$ training acceleration. A multi-objective scoring strategy further enhances robustness in compositional image generation. Inversion-DPO achieves SOTA performance on both text-to-image generation and compositional image generation. We additionally curate a structured dataset of 11,140 annotated images to support complex scene synthesis.


\section{Acknowledgement}

This paper is supported by Provincial Key Research and Development Plan of Zhejiang Province under No. 2024C01250(SD2) and National Natural Science Foundation of China (Grant No. 62006208).

\bibliographystyle{ACM-Reference-Format}
\bibliography{sample-base}

\appendix

\clearpage

%%%%%%%%%%%%%%%%%%%%%%%%SM%%%%%%%%%%%%%%%%%%%%%%%%%
\begin{center}
\Large \textbf{Supplementary Material}
\end{center}

\setcounter{section}{0}
\setcounter{table}{0}
\setcounter{figure}{0}
\setcounter{equation}{0}

\renewcommand{\thesection}{S\arabic{section}}
\renewcommand{\thetable}{S\arabic{table}}
\renewcommand{\thefigure}{S\arabic{figure}}
\renewcommand{\theequation}{S\arabic{equation}}

%前面主要讲fundamental，一系列task放到一两句话。在一系列任务验证，t2i和更复杂的任务。
%弱化LAION-SG，说造了一个pair的数据，不要提具体数据集，可以在实现细节讲一下用了LAION-SG
%SG不要讲，只在数据集实现细节的时候说一两句，其他都用structure
%intro（分两段）和method把题目两点强调说，实验也是从这两个点。
%引用全面，其他文章有的指标都放上

\section{User Study}
To evaluate the alignment between the generated image and human cognition, we design a user study involving two control groups: the group of SDXL~\cite{podell2023sdxlimprovinglatentdiffusion} vs. Inversion DPO and the group of Diffusion-DPO~\cite{Wallace2024DiffusionDPO} vs. Inversion-DPO. For each control group, we randomly select 500 pairs of images. In each trial, users are presented with two images and their corresponding prompts (text descriptions), one pair coming from SDXL and Inversion-DPO, and the other from Diffusion-DPO and Inversion-DPO. Users' task is to choose the image that best matched the corresponding prompt. It is important to note that some of the generated images contain inappropriate content, so we remove a few images to ensure the experiment adhered to ethical guidelines and to prevent any potential negative impact.

The study invite 10 participants, with a gender ratio of 1:1, and ages ranging from 20 to 30. Participants come from diverse backgrounds, including computer science, design, and human-computer interaction (HCI).

\begin{table*}[ht]
  \centering
  \begin{tabular}{@{}lcc@{}}
    \toprule
    \textbf{User Preference} & \textbf{Other Methods} & \textbf{Inversion-DPO} \\
    \midrule
    SDXL vs Inversion-DPO & 0.3347 & 0.6653   \\ 
    Diffusion-DPO vs Inversion-DPO & 0.2733 & 0.7267  \\ 
    \bottomrule
  \end{tabular}
  \caption{Comparison of user preference between different image generation methods. In the comparison, Inversion-DPO consistently shows a higher preference from users, indicating its better alignment with human cognition and image content.}
  \label{tab: user study}
\end{table*}

In the comparison between SDXL and Inversion-DPO, 66.53\% of users select the image generated by Inversion-DPO, while only 33.47\% chose the image generated by SDXL. In the comparison between Diffusion-DPO and Inversion-DPO, Inversion-DPO has a higher selection rate of 72.67\%, while Diffusion-DPO has only 27.33\%. Based on the results of these two control experiments, Inversion-DPO shows a clear user preference, whether compared with SDXL or Diffusion-DPO. Particularly in the comparison with Diffusion-DPO, the selection rate for Inversion-DPO was significantly higher, indicating that Inversion-DPO has a distinct advantage in terms of content alignment and consistency with human cognition when generating images.

\textbf{Human Subjects Notification.}Before the experiment began, we provided a notification to the participants to inform them about the collection and use of data.

Dear volunteers, thank you for your support of our research. We are studying an image generation algorithm that can generate high-quality images based on user-provided text prompts. All information about your participation in this study will be included in the study records. All information will be processed and stored according to the local privacy laws and policies. Your name will not appear in the final report. When referring to your data, only the individual number assigned to you will be mentioned.

 The use of user data has been approved by the Institutional Review Board of the primary author's affiliation.

\section{Additional Generation Examples}

\begin{figure*}[t]
    \centering
    \includegraphics[width=1\textwidth]{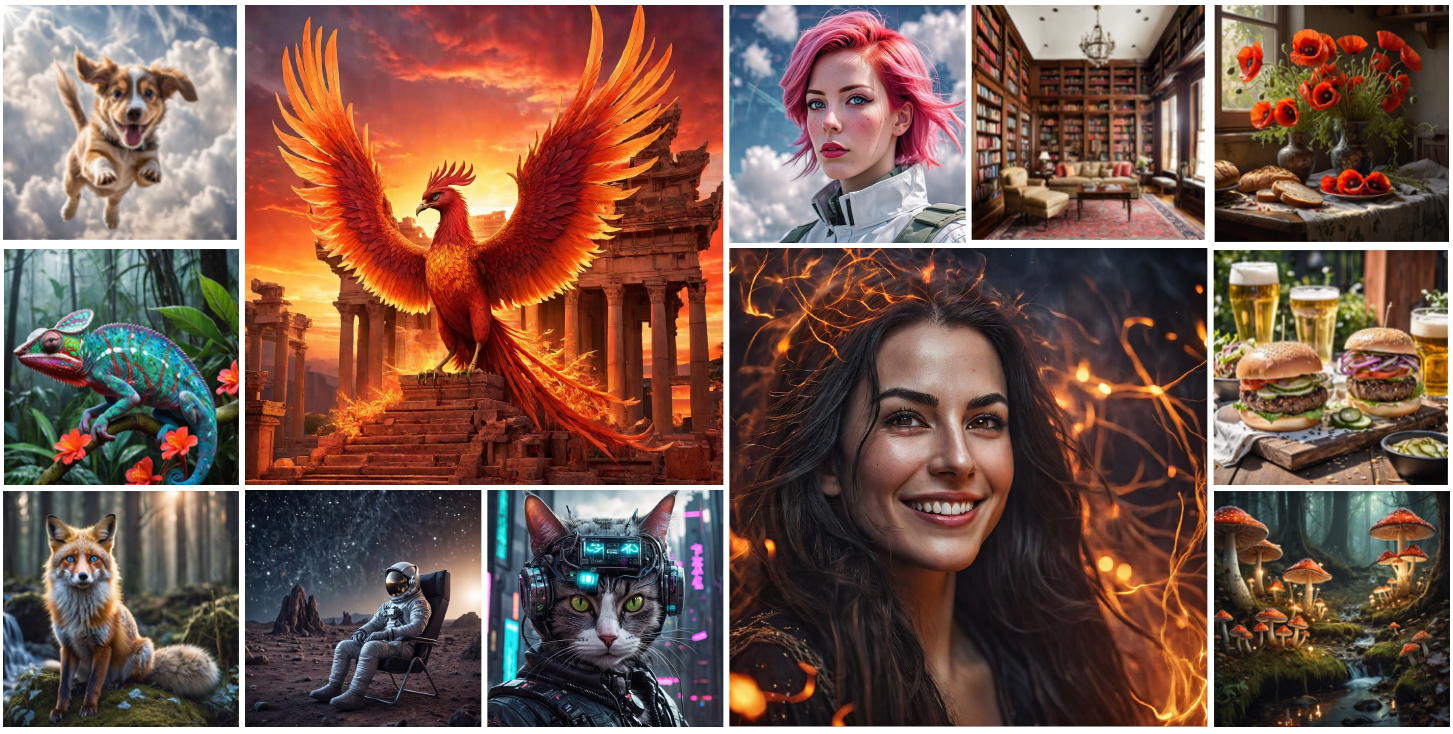} % Reduce the figure size so that it is slightly narrower than the column.
    % \vspace{-1.5em}
    \caption{More high-quality and visually appealing generation examples.}
    \label{fig: examples_successful}
    % \vspace{-1em}
\end{figure*}

\begin{figure*}[t]
    \centering
    \includegraphics[width=1\textwidth]{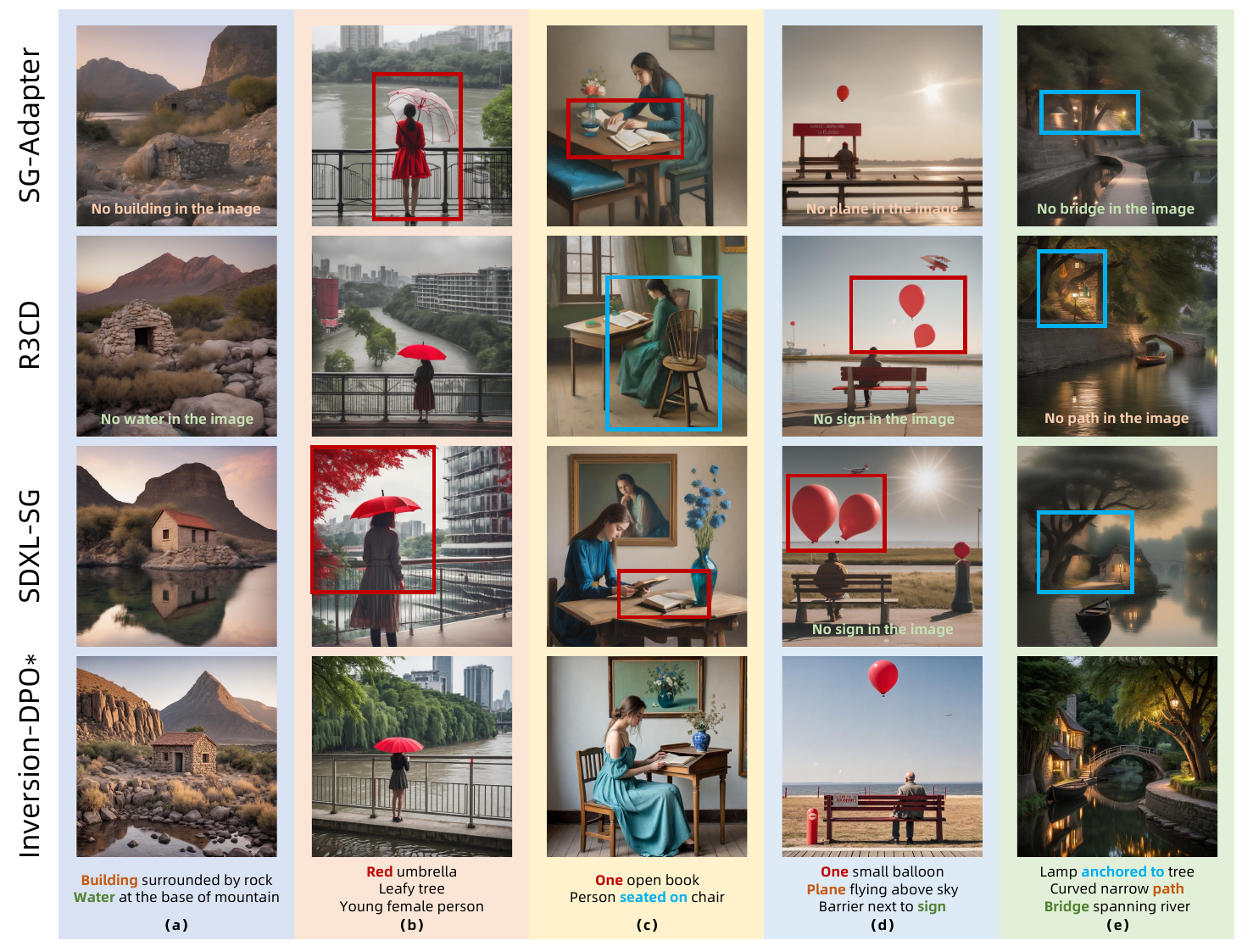} % Reduce the figure size so that it is slightly narrower than the column.
    % \vspace{-1.5em}
    \caption{Qualitative results on compositional image generation. Inversion-DPO proves effective in addressing key challenges in compositional image generation, including object omission (orange and green annotations in (a), (d), and (e)), attribute errors (red annotations in (b), (c), and (d)), and relation inaccuracies (blue annotations in (c) and (d)). As shown in the last row, Inversion-DPO achieves satisfactory generation results on cases where baseline models struggle.}
    % compositional image generation的quallitative result。Inversion-DPO在解决compositional image generation的现存问题上是有效的。这些问题包括：object缺失（（a）（d）（e）中的橙色和绿色标注）,属性错误（（b）(c)(d)中的红色标注），和关系错误（(c)（d）中的蓝色标注）。 baseline难以handle的问题Inversion-DPO都做到了令人满意的生成（最后一行）。}
    \label{fig: qualitative-sg2im}
    % \vspace{-1em}
\end{figure*}

% Inversion-DPO的生成图像能力达到了极高水平，图X展示了它生成的极具高质量的结果图像。同时，不可否认它也有生成失败结果的时候。比如生成的腿、翅膀和头部混乱，或者手指数量不对，如图X中显示的那样。

Inversion-DPO demonstrates a highly competent capability in image generation, achieving visually impressive and high-quality results, as illustrated in \cref{fig: examples_successful}. However, it is necessary to acknowledge that the model is not immune to failure cases. Typical failure modes include structural inconsistencies such as distorted legs, wings, or heads, as well as incorrect finger counts, as shown in the examples in \cref{fig: examples_failure}.

\section{Additional Explanation of Efficiency}

\begin{figure}[t]
    \centering
    \includegraphics[width=0.9\columnwidth]{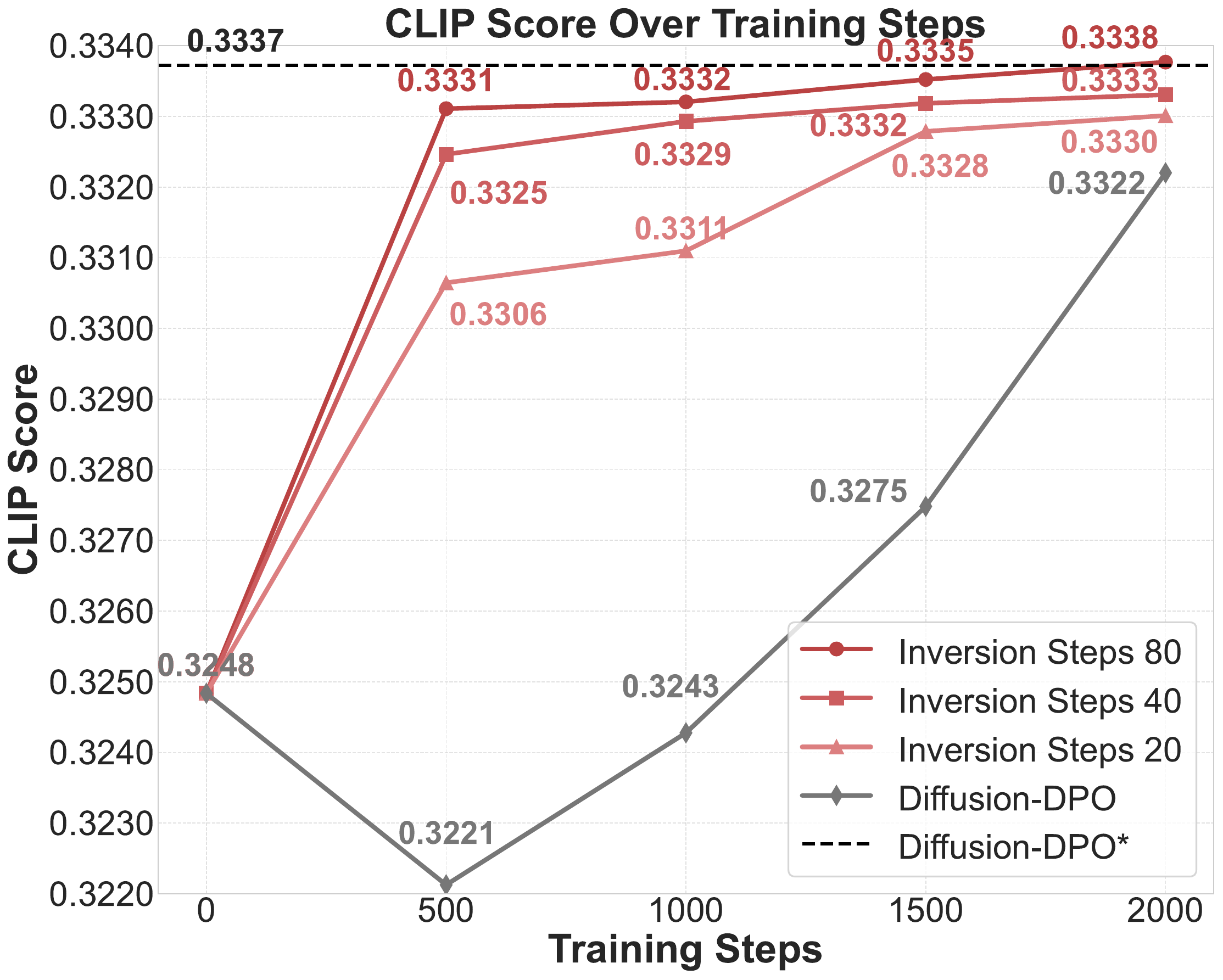} % Reduce the figure size so that it is slightly narrower than the column.
    \caption{Relationship between training steps and CLIP Score improvements. The red curve shows the performance of our Inversion-DPO trained with different numbers of DDIM inversion steps. The gray curve represents our implementation of Diffusion-DPO~\cite{Wallace2024DiffusionDPO}, while the dashed line corresponds to the officially released weights of Diffusion-DPO. All models are trained under identical configurations.}
    \label{fig: efficiency_clip_score}
\end{figure}

In addition to the relationship between training steps and PickScore~\cite{Yuval2023Pick-a-Pic}, as well as between training steps and Aesthetics Score~\cite{LAION_5B} presented in the main paper, we also report the correlations between training steps and CLIP Score~\cite{Radford2021Learning_CLIP} (\cref{fig: efficiency_clip_score}). Under identical training configurations, Inversion-DPO consistently outperforms Diffusion-DPO across all metrics, while also demonstrating faster convergence. Notably, even under constrained training conditions due to hardware limitations, Inversion-DPO still surpasses the officially released Diffusion-DPO in final performance.

% 除了main paper中展示的训练步数与PickScore之间的关系，我们同时report了训练步数和Aesthetics Score的关系与训练步数和CLIP Score的关系。当采用相同的训练配置时，Inversion-DPO在各项指标上的表现都超过了Diffusion-DPO，并且训练收敛速度更快。即便因硬件等原因Inversion-DPO带有训练限制，其最终表现仍超过官方发布的Diffusion-DPO.

\section{Visual Interpretation of the Underlying Assumptions}

\begin{figure*}[t]
    \centering
    \includegraphics[width=1\textwidth]{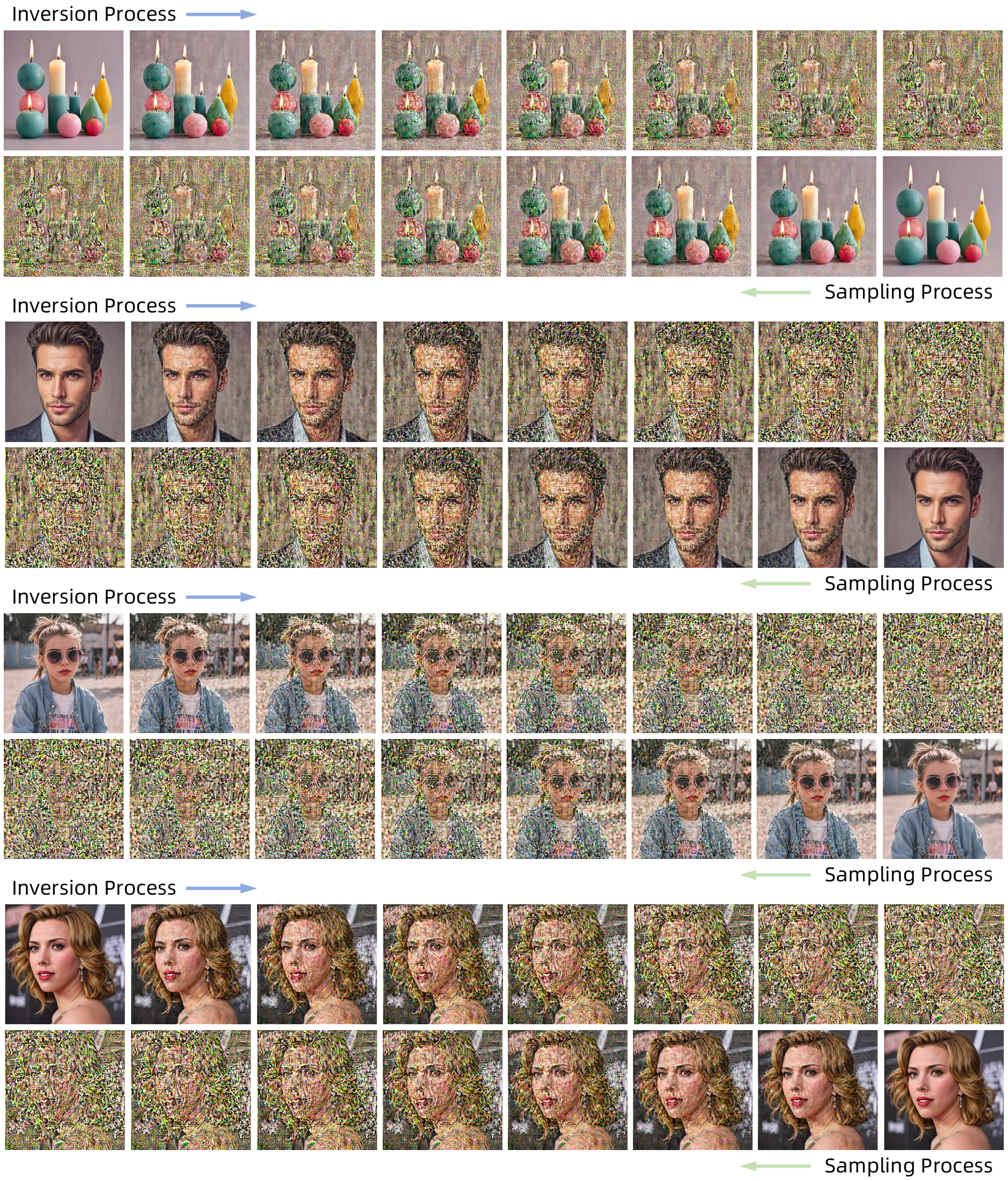} % Reduce the figure size so that it is slightly narrower than the column.
    % \vspace{-1.5em}
    \caption{Visualization of data distributions in the inversion and sampling processes.}
    \label{fig: inver_sample}
    % \vspace{-1em}
\end{figure*}

% 在我们推导中，有一个基于DDIM inversion的假设是the values of $\boldsymbol{x}_{t-1}^w$ are similar in the inversion from $x_0^w$ and in the sampling from $x_T^w$. 为验证这一假设的可靠性，我们对inversion轨迹和sampling轨迹的分布情况进行了实时可视化。如图X中所示，可以看到不论处于正向过程还是逆向过程，对应的时间步骤的数据的视觉呈现是近似一致的，which直观的验证了这一假设的可靠性。

In our derivation, a key assumption based on DDIM inversion~\cite{song2020ddim} is that the values of \(\boldsymbol{x}_{t-1}^w\) are similar when obtained from the inversion of \(x_0^w\) and from the forward sampling of \(x_T^w\). To assess the validity of this assumption, we perform real-time visualization of the distributions along both the inversion and sampling trajectories. As illustrated in \cref{fig: inver_sample}, the visual representations at corresponding time steps exhibit high similarity in both the forward and reverse processes, providing intuitive evidence that supports the reliability of this assumption.

\section{Detailed Derivation of Inversion-DPO}
We present detailed derivation of the final loss of InversionDPO. We begin with the form as
\begin{equation}
\log\frac{p_\theta(\boldsymbol{x}_{0:T}^w)}{p_{\theta_0}(\boldsymbol{x}_{0:T}^w)}
=
\sum_{t=1}^T\left[\log p_\theta(\boldsymbol{x}_{t-1}^w|\boldsymbol{x}_t^w)-\log p_{\theta_0}(\boldsymbol{x}_{t-1}^w|\boldsymbol{x}_t^w)\right]
\end{equation}

Given the $\boldsymbol{x}_{1:T}^w$ by DDIM Inversion, we derive the following computation within the expectation.
Specifically, $\log\frac{p_\theta(\boldsymbol{x}_{0:T}^w)}{p_{\theta_0}(\boldsymbol{x}_{0:T}^w)}$ is first decomposed into the differences in conditional probabilities at each time step:
\begin{equation}
\log\frac{p_\theta(\boldsymbol{x}_{0:T}^w)}{p_{\theta_0}(\boldsymbol{x}_{0:T}^w)}
=
\sum_{t=1}^T\left[\log p_\theta(\boldsymbol{x}_{t-1}^w|\boldsymbol{x}_t^w)-\log p_{\theta_0}(\boldsymbol{x}_{t-1}^w|\boldsymbol{x}_t^w)\right]
\end{equation}
% 因x和y的a分布相同，故抵消。
Since \( p_{\theta} \) and \( p_{\theta_0} \) share the same distribution of \( x_T \) as standard Gaussian, $\log p_{\theta}(x_T^w) - \log p_{\theta_0}(x_T^w)$ cancel out. 
% 使用$p_\theta$的DDIM inversion之后得到的 \((x_1, \dots, x_T)\)，其中给定$x_t$，根据公式（7）和（9），每一个采样出来的$x_{t-1}$都是确定性的，可以写成：
% After applying DDIM inversion under \( p_\theta \), we obtain the sequence \( (x_1^w, \dots, x_T^w) \).
The conditional probabilities of \( p_{\theta} \) and \( p_{\theta_0} \) are Gaussian distributions as in equation~8 and becomes dirac-delta as $\sigma_t$ approaches $0$.

Given the assumption of inversion in Sec.~3.2, the values of $\boldsymbol{x}_{t-1}^w$ are similar in the inversion from $x_0^w$ and in the sampling from $x_T^w$. 
Therefore, $\log p_\theta(\boldsymbol{x}_{t-1}^w|\boldsymbol{x}_t^w)$ is only related to $\sigma_t^2$ and thus constant. 
At the same time, we have for \( p_{\theta_0} \) its expecation is
\begin{align}
    &\boldsymbol{\mu}_{\theta_0}(x_t^w) = \frac{\sqrt{\alpha_{t - 1}}}{\sqrt{\alpha_t}}\boldsymbol{x}_t^w + \\
    &\left(\sqrt{1 - \alpha_{t - 1} - \sigma_t^2}-\frac{\sqrt{\alpha_{t - 1}({1 - \alpha_t})}}{\sqrt{\alpha_t}}\right)\epsilon_{\theta_0}(\boldsymbol{x}_t^w, t)
\end{align}

Therefore, we have 

\begin{equation}
    \begin{aligned}
        -\log p_{\theta_0}(\boldsymbol{x}_{t-1}^w|\boldsymbol{x}_t^w) &=  \frac{\|\boldsymbol{\mu}_{\theta_0}(x_t^w)-\boldsymbol{x}_{t-1}^w\|}{2\sigma_t^2}^2 +C\\
    \end{aligned}
\end{equation}

Since \( x_{t-1}^w \) is equivalently generated by \( p_{\theta} \) (i.e., \( \mu_\theta \)) based on the assumption above, we further assume \( x_{t-1}^w \) subjects to $\mathcal{N}( \mu_\theta(x_t^w), \sigma_t^2I)$. By
substituting it, we have the simplified form $\frac{1}{2\sigma_t^2}\|\boldsymbol{\mu}_\theta-\boldsymbol{\mu}_{\theta_0}\|^2$. 
% 将均值差异转换为epsilon预测的差异，设均值形式为a，得到：
Transforming the mean difference into the difference in epsilon prediction, and denoting the mean as $\boldsymbol{\mu}_\theta=\frac{1}{\sqrt{\alpha_t}}\boldsymbol{x}_t^w-\frac{\sqrt{1-\alpha_t}}{\sqrt{\alpha_t}}\epsilon_\theta(\boldsymbol{x}_t^w,t)$, we obtain:
\begin{equation}\|\boldsymbol{\mu}_\theta-\boldsymbol{\mu}_{\theta_0}\|^2=\frac{1-\alpha_t}{\alpha_t}\|\epsilon_\theta(\boldsymbol{x}_t^w,t)-\epsilon_{\theta_0}(\boldsymbol{x}_t^w,t)\|^2\end{equation}
% 将所有时间步的贡献相加，得到最终目标函数：
Summing the contributions across all time steps, we can arrive at:
\begin{equation}\log\frac{p_\theta(\boldsymbol{x}_{0:T}^w)}{p_{\theta_0}(\boldsymbol{x}_{0:T}^w)}=\sum_{t=1}^T\frac{1-\alpha_t}{2\sigma_t^2\alpha_t}\|\epsilon_\theta(\boldsymbol{x}_t^w,t)-\epsilon_{\theta_0}(\boldsymbol{x}_t^w,t)\|^2, \end{equation}
We adopt the simple form as in DDPM~\cite{ddpm} to eliminate the weighs and coefficients. Therefore, taking the winning and lossing sample all together, we reach the final loss as 
\begin{equation}\begin{aligned}
 \mathcal{L}_{\text{Inversion-DPO}}(\theta)=-\mathbb{E}_{(\boldsymbol{x}_0^w,\boldsymbol{x}_0^l)}\log\sigma\bigg(  \beta\mathbb{E}_{\substack{ \boldsymbol{x}_{1:T}^w\sim p_\theta(\boldsymbol{x}_{1:T}^w|\boldsymbol{x}_0^w) \\ 
 \boldsymbol{x}_{1:T}^l\sim p_\theta(\boldsymbol{x}_{1:T}^l|\boldsymbol{x}_0^l)}}\\
\sum_{t=1}^T\Bigg[ \|\epsilon_\theta(\boldsymbol{x}_t^w,t)-\epsilon_{\theta_0}(\boldsymbol{x}_t^w,t)\|^2- \|\epsilon_\theta(\boldsymbol{x}_t^l,t)-\epsilon_{\theta_0}(\boldsymbol{x}_t^l,t)\|^2\Bigg]\bigg)
\end{aligned}\end{equation}

\section{Impact of the Imprecise Approximation}

\begin{table}
    \centering
    \begin{tabular}{@{}lccc@{}}
    \toprule
        \textbf{Model} & \textbf{PickScore}$^\uparrow$ & \textbf{CLIP}$^\uparrow$ & \textbf{Aesthetic}$^\uparrow$ \\
        \midrule
        Diffusion-DPO (\( q \)) & 0.227 & 0.340 & 6.05  \\ 
        Inversion-DPO (\( p_\theta \)) & 0.228 & 0.341 & 6.11  \\ 
        Diffusion-DPO$\mathcal{y}$ (\( q \)) & 0.212 & 0.324 & 5.73  \\ 
        Inversion-DPO$\mathcal{y}$ (\( p_\theta \)) & 0.212 & 0.325 & 5.73  \\ 
        \bottomrule
    \end{tabular}
    % \vspace{-0.5em}
    \caption{Performance impact of approximating the reverse distribution \( p_\theta \) with the forward \( q \). In Inversion-DPO, the true reverse distribution \( p_\theta \) is utilized, whereas Diffusion-DPO relies on its approximation \( q \).}
    \label{tab: ablation-pq}
    % \vspace{-1em}
\end{table}

In Diffusion-DPO, due to the inaccessibility of the reverse trajectory, the posterior sampling \( p_\theta(x_{1:T} | x_0) \) is approximated by the noisy process \( q(x_{1:T} | x_0) \). We argue that this approximation introduces non-negligible training errors and conduct an ablation study to investigate this effect. While Inversion-DPO involves recovering the whole noise trajectory across multiple time steps for alignment, in the code implementation, the original Diffusion-DPO samples a single timestep from \( q(x_{1:T} | x_0) \) and augment with a value $T$ to obtain the final result. 
Such difference may also introduce performance discrepancy.
For a fair comparison, we also evaluate Inversion-DPO by randomly sampling a single timestep from \( p_\theta(x_{1:T} | x_0) \). 
As presented in \cref{tab: ablation-pq} in SDXL and SD1.5, training with DDIM Inversion with \( p_\theta \) consistently yields better results than approximation based on the forward distribution \( q \), highlighting the necessity and effectiveness of the core insight behind Inversion-DPO.
% 不论backbone采用SDXL还是SD1.5，采用真实reverse distribution p_\theta训练得到的结果总是比采用forward distribution q拟合的要好，说明了Inversion-DPO的insight的必要性。

% 我们也分析了不同的DDIM Inversion步数对模型训练的影响，结果如表X。it is明显的that随着Inversion steps的增加，模型在各个方面的表现性能越来越好。这是由于当DDIM Inversion的时间步较多时，相邻step之间的分布差异越小，从而更容易引导模型向正确的方向优化。最终Inversion步数为80的配置被选择去进行Inversion-DPO的训练。
\section{A Brief Introduction of Stable Diffusion}

Diffusion models, such as Denoising Diffusion Probabilistic Models (DDPMs)~\cite{ddpm} and Denoising Diffusion Implicit Models (DDIMs)~\cite{song2020ddim}, provide a generative modeling framework that is based on the gradual corruption of data through a noise injection process followed by the recovery of the original data. These models leverage a Markov chain process to progressively add Gaussian noise to the data in a forward process, and then learn to reverse this process to recover the data from noisy samples.

In DDPM, the forward process is a Markov chain where data \(x_0\) is progressively perturbed by Gaussian noise at each timestep, with the perturbation increasing over time. The transition between timesteps \(x_{t-1}\) and \(x_t\) is governed by a Gaussian distribution:

\[
q(x_t | x_{t-1}) = \mathcal{N}(x_t; \sqrt{1 - \beta_t} x_{t-1}, \beta_t I),
\]

where \(\beta_t\) is the noise variance at timestep \(t\), and the forward process gradually transforms the data \(x_0\) into pure noise \(x_T\). The reverse process, which is modeled during training, aims to recover the data by learning to reverse the noisy transformations:

\[
p_\theta(x_{t-1} | x_t) = \mathcal{N}(x_{t-1}; \mu_\theta(x_t, t), \Sigma_\theta(x_t, t)).
\]

The model is trained by optimizing the evidence lower bound (ELBO), which involves minimizing the difference between the forward and reverse process distributions. This training objective can be written as:

\[
\mathcal{L}_{\text{simple}} = \mathbb{E}_{x_0, \epsilon, t} \left[ \left\| \epsilon - \epsilon_\theta(\sqrt{\bar{\alpha}_t} x_0 + \sqrt{1 - \bar{\alpha}_t} \epsilon, t) \right\|^2 \right],
\]

where \(\epsilon_\theta\) is the model’s prediction of the noise added at each timestep, and \(\epsilon\) is Gaussian noise. This loss function allows the model to learn to predict the noise at each timestep, effectively learning the reverse process.

To accelerate the sampling process, \cite{song2020ddim} introduces DDIMs, which extend DDPMs by generalizing the forward process to be non-Markovian. This modification allows for a more efficient sampling process that reduces the number of steps required to generate high-quality samples. In DDIMs, the generative process is deterministic, and the model can sample from fewer timesteps while maintaining high sample quality. The key insight behind DDIMs is that the reverse process in DDPM can be reparameterized using a deterministic trajectory, which simplifies the sampling process significantly.

\section{Limitation}

\begin{figure*}[t]
    \centering
    \includegraphics[width=1\textwidth]{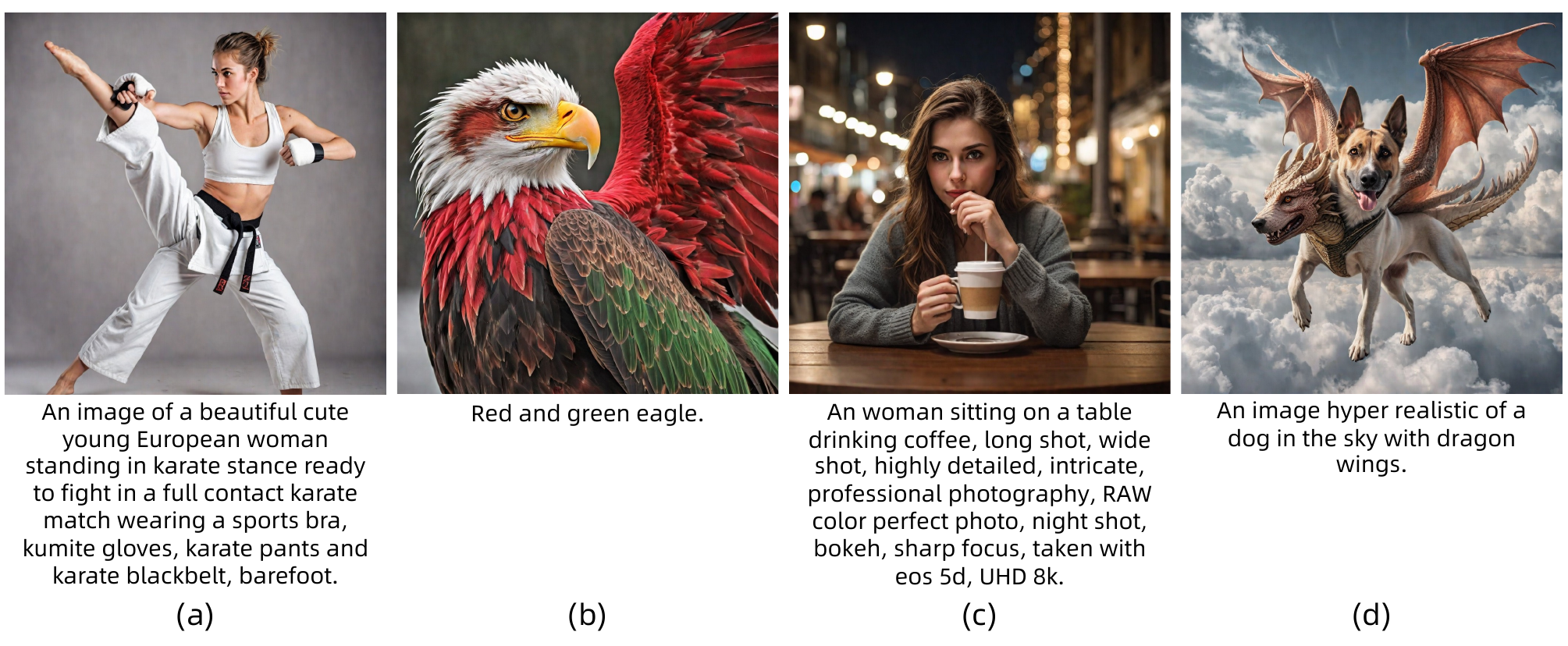} % Reduce the figure size so that it is slightly narrower than the column.
    % \vspace{-1.5em}
    \caption{Additional results of failure examples.}
    \label{fig: examples_failure}
    % \vspace{-1em}
\end{figure*}

Despite the effectiveness of our proposed Inversion-DPO framework, several limitations remain. First, the DDIM Inversion process relies on a strong assumption that the noise distribution in the forward (diffusion) and reverse (denoising) processes is consistent. 
Such assumption is valid only when the number of inversion steps is large enough.
In practice, with a small inversion step, this assumption may potentially introduce inaccuracies in the reconstructed trajectories. 
Second, the training pipeline requires two rounds of inference: one from the pretrained model \( p_{\theta_0} \) to obtain the DDIM-inverted noise trajectories, and another from the reference model \( p_\theta \) to estimate the output distribution. This two inferences improve the convergence but increase training time and computational cost. Finally, the exact impact of the errors introduced by DDIM inversion remains unclear, and their influence on the overall optimization process warrants further investigation. More accurate Inversion methods may lead to more precision approximation.
We present some failure cases in \cref{fig: examples_failure}. Such failure may be a result of insufficient training or inadequate training data for aligment.

\section{Social Impact}

Our work enhances preference optimization in diffusion models, offering improved controllability and sample quality. However, this also introduces potential risks. Automated preference signals, if not carefully designed, may inadvertently promote unsafe or biased content due to metric misalignment. Moreover, there exists the possibility of malicious misuse, where adversarial inputs could manipulate the optimization to favor harmful or misleading generations. We advocate for incorporating rigorous safety filters, human oversight, and transparent evaluation practices to mitigate such risks and ensure responsible deployment.

\end{document}